\documentclass[conference]{IEEEtran}

\ifCLASSINFOpdf
\else
\fi
\usepackage{url}
\usepackage{amsmath}
\usepackage{graphicx}
\usepackage{multicol}
\usepackage{comment}
\usepackage{multirow} 
\usepackage{epstopdf}
\usepackage{color}
\usepackage{diagbox}
\usepackage{latexsym} 
\usepackage{times}
\usepackage{graphicx}
\usepackage{booktabs}
\usepackage{graphicx}
\usepackage{subfigure}
\usepackage{float}
\usepackage{mathtools, amsfonts, amssymb, amsthm,color,bm} 
\usepackage{cite}
 \usepackage[english]{babel}
\usepackage[figurename=Fig.,font={small,it}]{caption} 
\usepackage{siunitx}
\usepackage{algorithm}
\usepackage{algorithmic} 

\usepackage[]{fancyhdr} %
\newcommand{\changefont}{\fontsize{9}{9}\selectfont}
\IEEEoverridecommandlockouts
\begin{document}

%
\title{Recent Results of Energy Disaggregation with Behind-the-Meter Solar Generation}

\author{\IEEEauthorblockN{ Ming Yi,  Meng Wang}
\IEEEauthorblockA{Department of Electrical, Computer, and
Systems Engineering, Rensselaer Polytechnic Institute\\
Troy, USA\\
}
 }


%





\maketitle
\thispagestyle{fancy}
\pagestyle{fancy}


\begin{abstract}

The rapid deployment of renewable generations such as photovoltaic (PV) generations brings great challenges to the resiliency of existing power systems. Because    PV  generations  are volatile and   typically invisible to the power system operator, estimating the generation and characterizing the uncertainty  are in urgent need for operators to make insightful decisions. This paper summarizes our recent results on energy disaggregation at the substation level with Behind-the-Meter solar generation.

We formulate the so-called ``partial label'' problem for energy disaggregation at substations, where the aggregate measurements contain the total consumption of multiple loads, and the existence of some loads is unknown. We develop two model-free disaggregation approaches based on deterministic dictionary learning and Bayesian dictionary learning, respectively. Unlike conventional methods which require fully annotated training data of individual loads, our approaches can extract load patterns given partially
labeled aggregate data. Therefore, our partial label formulation is more applicable in the real world.  Compared with deterministic dictionary learning, 
the Bayesian dictionary learning-based approach provides the uncertainty measure for the disaggregation results, at the cost of increased computational complexity. All the methods are validated by numerical experiments.
\end{abstract}



%
\IEEEpeerreviewmaketitle

\section{Introduction}

{The  presence of renewable generations in power systems, especially solar generations, has increased rapidly in recent decades. Reference \cite{SolarMart} reports that the global capacity of photovoltaic (PV) installment reached 634 GW in 2019.  The solar capacity in 2019 has grown nearly 400 times since 2000. California Independent System Operator (ISO) estimates that the renewable energy generations will contribute 50\% power supplies by 2030 in California \cite{CAISO}.  

The wide deployment of renewable generations 
decreases greenhouse gas emissions, however, but also brings great challenges to the reliability and resiliency of existing power systems. For example, at the substation level, the measurements of power consumptions are the net loads that contain different types of loads.  The solar generation is invisible to the power system operator and thus  is behind-the-meter (BTM). {Because of the stochastic nature and high volatility of renewable generations, the accurate estimation of generated energy is challenging.}
Energy disaggregation at the substation level (EDS) 
aims to disaggregate each individual load\footnote{{Generation is considered as a negative load in this paper.}}  from aggregate measurement. The accurate information for load consumption are crucial for power system planning and operations,  such as hosting capacity evaluation \cite{AMZ18,WDW19}, demand response and load dispatching \cite{MB20,XGL20} and load forecasting \cite{WZCK18,SZWS19}.
 
 {Energy disaggregation problem at the household level (EDH) has been extensively studied, see, e.g., 
\cite{KBNY10,KDM16,CZWHFJ19,HSLS16}, 
also under the terminology \textit{non-intrusive load monitoring (NILM)} \cite{KBNY10,KDM16,CZWHFJ19,HSLS16,H92,GAM15,MSW16}}. The electric appliances are typically single-state or multi-state devices and patterns of their power consumptions usually are repeatable. The general procedure for EDH methods is to first collect  historical power consumption for each individual appliance and learn patterns from these well-annotated data. Then EDH methods disaggregate power consumptions for each appliance from the aggregate data based on these patterns. In comparison, obtaining historical power consumption for each individual load at the substation level is more difficult, as the measurements at the substation level are highly aggregated from different types of loads. Even though the operator has information about load types attached to a substation,  whether a   certain load is consuming/generating energy or not in a certain time interval is not already clear. One example is the BTM solar generation. 
{Thus, the measurements at the substation  usually   contain multiple loads and are partially labeled.  
 It is more challenging to learn distinctive load profiles 
under this situation than learning from measurements on individual loads. Moreover, the volatility of load and renewable generation often lead to significant estimation errors. }
However, to the best of our knowledge, there is no work that provides a confidence measure of the energy disaggregation results.

{ This paper summarizes our recent results
for solving these two challenges. Given partially labeled training data,  our work~\cite{LYWW20} proposes a  deterministic dictionary learning-based method to learn load patterns and disaggregate the aggregate measurements into individual loads in real-time.
Note that \cite{LYWW20} is a deterministic approach and therefore is unable to provide the confidence measure of the estimation results. To estimate the reliability of the disaggregation results, 
in \cite{YW21}, we propose a probabilistic energy disaggregation approach based on Bayesian dictionary learning.  }

{The contributions of this paper are three folds: 1. We summarize our works~\cite{LYWW20}  and  \cite{YW21} for solving the ``partial label'' problem and modeling the uncertainty. 2. We compare  these two methods and other two existing works in the experiment. (3) We provide  more testing cases for these two methods in this paper. }

{The remainder of this paper is organized as follows. Section II explains our partial label formulation. Section
III discusses our proposed deterministic approach to solve the issue of  partial labels and introduces our proposed Bayesian method for modeling the uncertainty of disaggregation results. Section IV summarizes this paper.}

\section{Problem Formulation}

{A substation is connected to  $C$ ($C>1$) types of loads in total. 
Let $\bm{x}\in \mathbb{R}^P$ denote the aggregate measurement with window length $P$. Let a binary vector $\bm{y}= [y^1, y^2,..., y^C] \subseteq \{0,1\}^C$ denote the  load existence in $\bm{x}$.  For example,
when $C=3$, $\bm{y}=[0,0,1]^T$ means that only load 3 exists in $\bm{x}$.}

{In our paper  \cite{LYWW20}, we propose a ``partial label formulation'' where the operator only knows partial entries in  $\bm{y}$. The partial labels can be obtained by designing a load detector for each load separately \cite{HCQ19,ZG16} or from engineering experience. As described in \cite{LYWW20}, annotating partial labels  has a lower cost for manpower or communication burdens than annotating all the labels. Moreover,  if a detector fails to identify some loads \cite{HD18}, we can only obtain partial labels.

Let $\bar{\bm{X}}=[\bar{\bm{x}}_1,\bar{\bm{x}}_2,...,\bar{\bm{x}}_N ] \in \mathbb{R}^{P \times N}$ denote  $N$ measurements.  $\bar{\bm{x}}_i$ denotes the data at the $i$th time window.  $\bm{y}_i \in \{0,1\}^C$ denotes the labels in $\bar{\bm{x}}_i$. Let label matrix $\bm{Y}=[\bm{y}_1, \bm{y}_2,...,\bm{y}_N]$ denote all the labels in $\bar{\bm{X}}$. Let $\Omega$ denote the indices of known entries in $\bm{Y}$. $\bm{Y}_\Omega$ denotes all the known partial labels. In the above example, if one only knows $\bar{\bm{x}}$ contains load 3 and does not know whether the other two loads exist or not, then the corresponding $\bm{Y}_\Omega$ is $[?, ?, 1]^T$ where $?$ denotes one does not know the corresponding load exists or not.

Fig. 1 illustrates our partial label formulation. The aggregate data are aggregated from  two industrial loads and one solar generation. Each subfigure shows patterns of aggregate data and the corresponding individual loads at the same time interval. {In all these four cases, the label is $[?, ?, 1]^T$, indicating that load 3 always exists, while the existence of loads 1 and 2 is unknown. }

 Given training dataset $\bar{\bm{X}}$, the corresponding partial label matrix $\bm{Y}_\Omega$  and an aggregate measurement $\hat{\bm{x}}\in \mathbb{R}^P$, the objective of this paper is to: (1)   learn distinctive patterns of individual loads from $\bar{\bm{X}}$ and disaggregate 
 $\hat{\bm{x}}$, and (2)   characterize the uncertainty of the disaggregation results.

}

    \begin{figure}[!ht]  
     \centering
    	\includegraphics[width=0.45\textwidth]{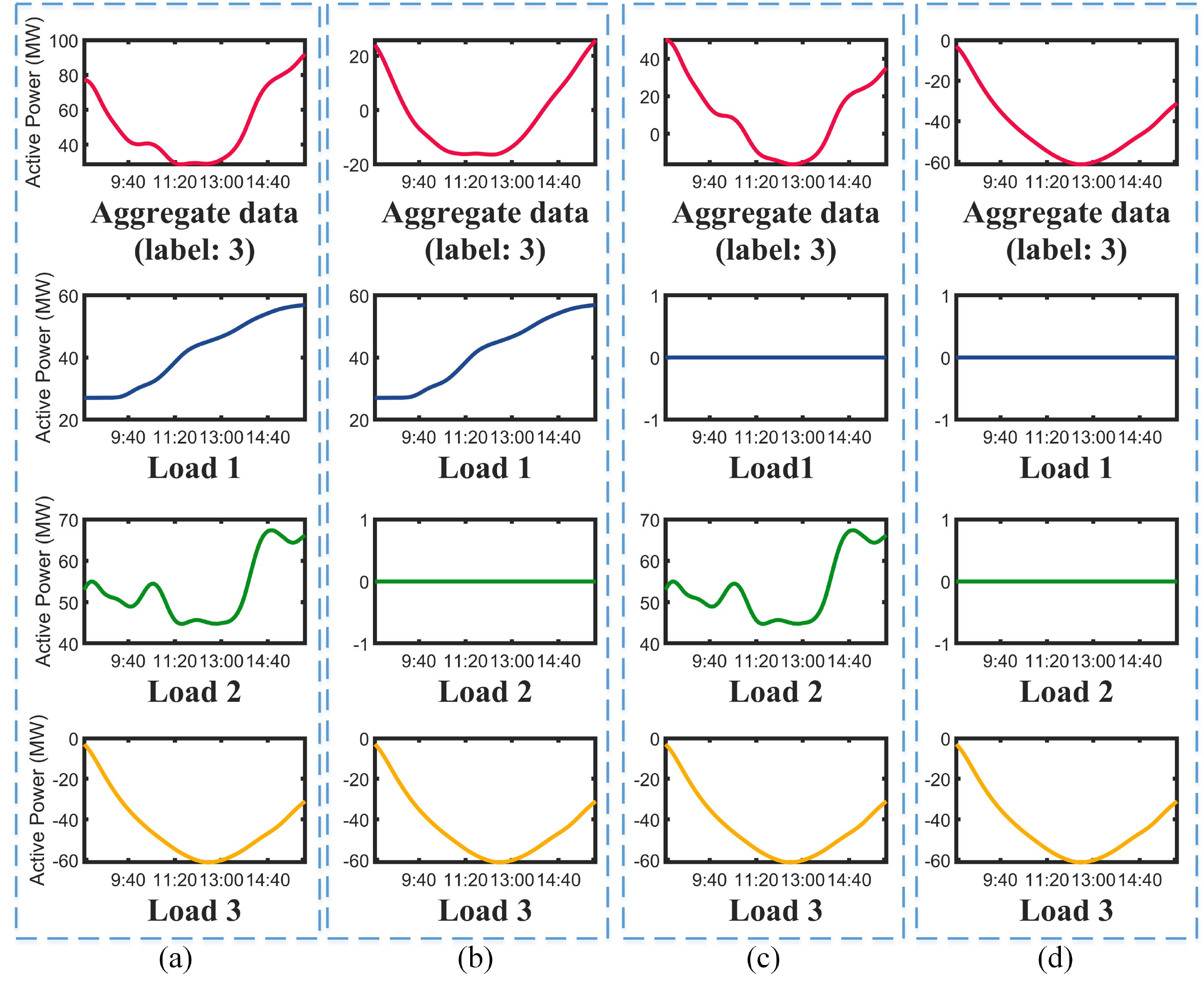}  
    	\caption{{An example of partial label formulation. There are three load types in the aggregate data. All four aggregate data have the same partial label  $[?,?,1]^T$.   The aggregate data contain (a) all three loads; (b) load 1 and 3; (c) load 2 and 3; (d) only load 3.} \cite{LYWW20}}  \label{fig1}
    \end{figure}

\section{Methodology}

{In this section, we present our two model-free approaches based on deterministic dictionary learning \cite{LYWW20} and Bayesian dictionary learning \cite{YW21}, respectively. 
\subsection{Deterministic Energy Disaggregation}

To learn patterns of each individual load from the given training data $\bar{X}$, we formulate 
a deterministic dictionary learning problem,
 \begin{align} 
\min_{ A, D }    f(A,D) &=\lVert \bar{X}- \Sigma_{i = 1}^C D_i A_i \rVert_F^2+\Sigma_{i = 1}^C \lambda_i \Sigma_{j: i \notin y_j }  \lVert A_i^{j} \rVert   \nonumber\\
 &   + \lambda_D \text{Tr}(D\Theta D^\intercal)    \label{eqn:dictionary} \\
\text{s.t. }&   \lVert d^{m } \rVert_2 \leq 1,   d^{m} \geq 0, m=1, \cdots, K     \label{cons} \\
& c_i A_i   \geq 0, \forall i   \label{eqn:positive}  
\end{align} 
where $D_i\in \mathbb{R}^{P\times K_i}$ denotes the dictionary for  load $i$, and $A_i   \in \mathbb{R}^{K_i \times N}$ denotes the corresponding coefficients   of load $i$.  $A_i^j$ is the $j$th column in $A_i$.   $A=[A_1; A_2; \cdots; A_C]\in \mathbb{R}^{K \times N}$ is the matrix that contains all coefficients. $D = \begin{bmatrix}
D_1, D_2, \cdots, D_C
\end{bmatrix}  \in \mathbb{R}^{P \times K }$    is the matrix that contains all dictionaries. $d^m$ is the $m$th column in $D$.  $K = \Sigma_{i=1}^C K_i $. $\text{Tr}(\cdot)$  is the trace operator, and $D^\intercal$ represents the transpose matrix of $D$.  $\lambda_i$ and $\lambda_D$ are pre-defined hyper-parameters.     }

   \begin{figure}[!ht]  
     \centering
    	\includegraphics[width=0.48\textwidth]{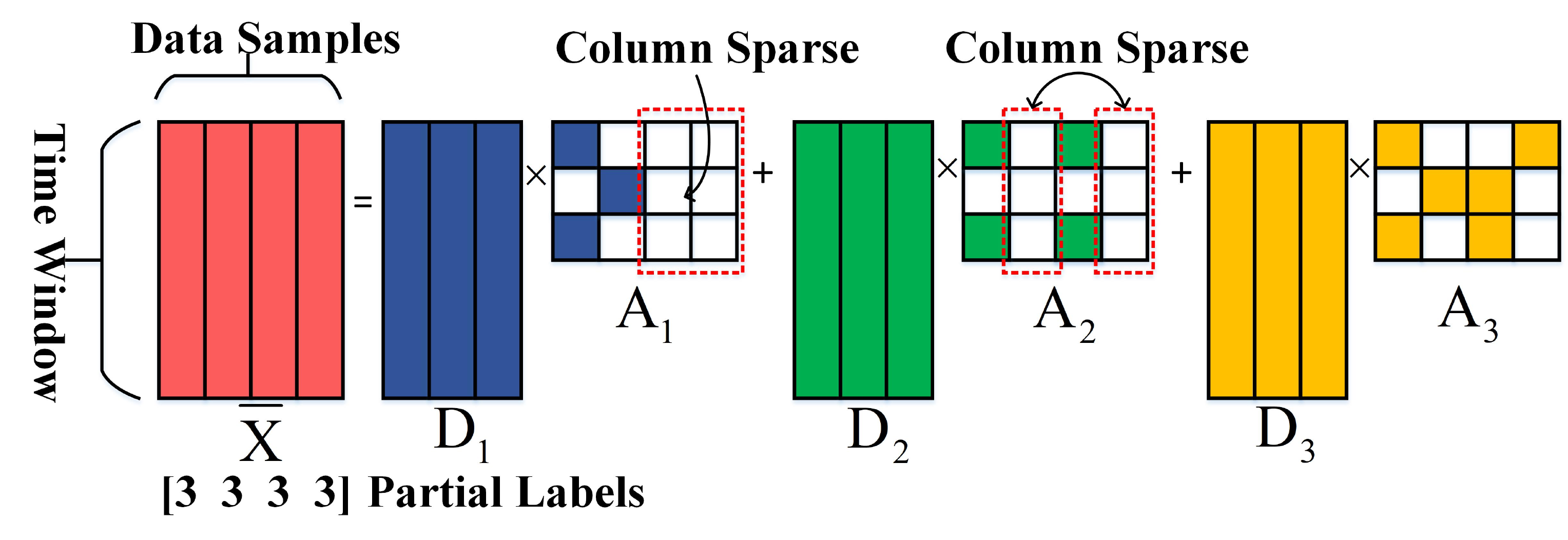}  
    	\caption{ {The dictionary representation in Fig. \ref{fig1}. 
     The coefficients	$A_1$ and $A_2$ are column-sparse.  \cite{LYWW20}}} 
    	\label{diction}
    \end{figure}

        \begin{figure*}[!ht]  
     \centering
    	\includegraphics[width=0.8\textwidth]{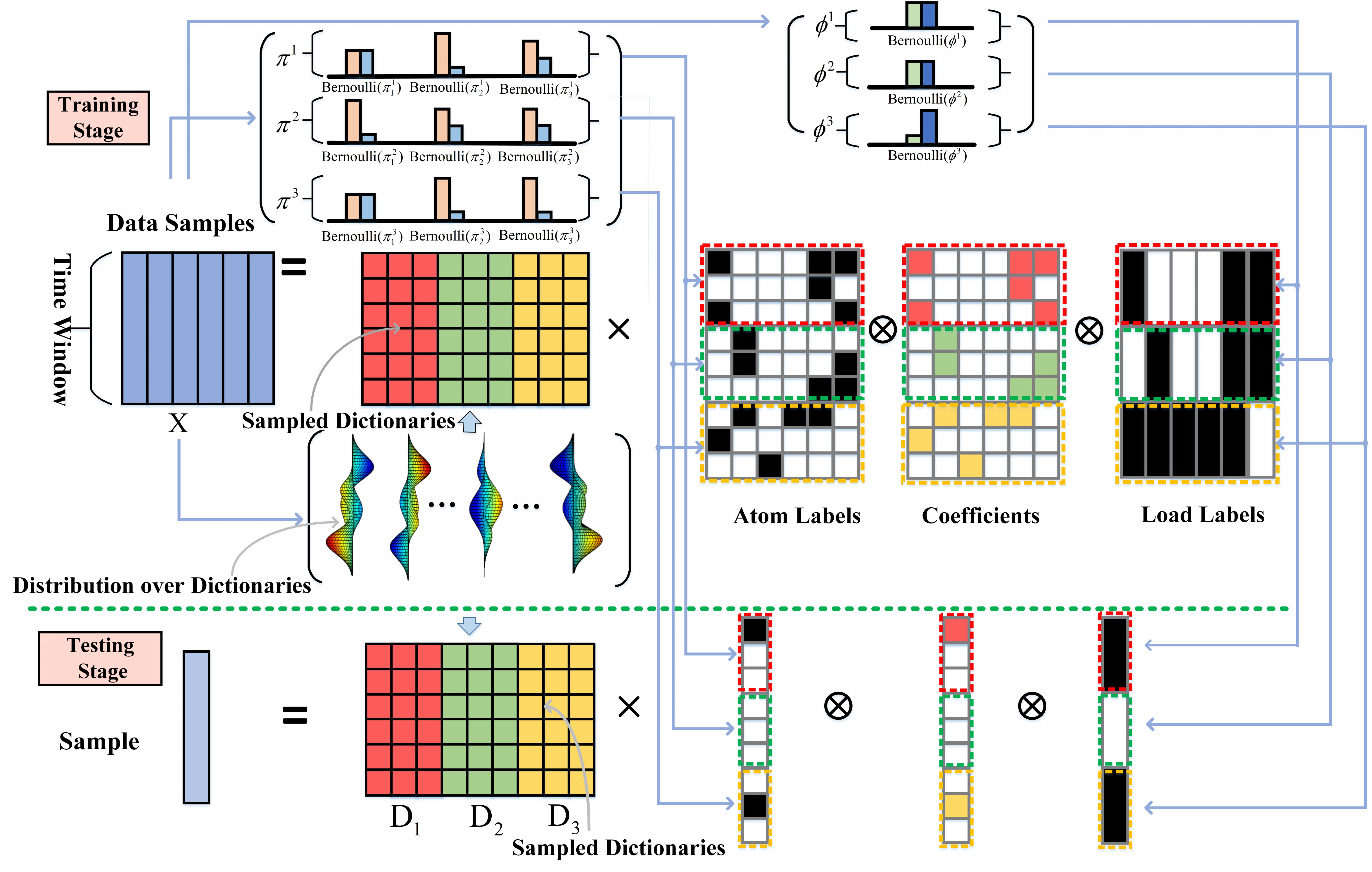}  
  \caption{{An illustrative framework of the proposed Bayesian method. In the training stage, our method learns the posterior distribution of atom labels, coefficients and load labels from the training data. In the testing stage, the method samples learned distributions of the dictionary and learns the posterior distribution for atom labels, coefficients and load labels, respectively, for test data. \cite{YW21}}}
    	\label{fig3}
  
  \label{fig_example}
    \end{figure*}
    
{The first term $\lVert \bar{X}- \Sigma_{i = 1}^C D_i A_i \rVert_F^2$ is the standard reconstruction error in dictionary learning. It measures the reconstruction error between the original data and the learned dictionaries and coefficients. $\sum_{j: i \notin y_j}  \lVert A_i^{j} \rVert$ is the column sparsity constraint. The motivation of using this regularization is illustrated in Fig. \ref{diction}, which shows the dictionary representation of Fig. \ref{fig1}. 
Because the training data only have partial label 3, load 1 and load 2 may not exist in the training data. Therefore, we impose the column sparsity on $A_1$ and $A_2$ to promote the group sparsity of $A_1$ and $A_2$. 
    
The incoherence term $\text{Tr} (D\Theta D^\intercal) $ is defined as

\begin{align}\label{eqn:trace}
\text{Tr} (D\Theta D^\intercal) & = \Sigma_{m = 1}^{K} \Sigma_{p =1}^{K } \theta_{mp}  (d^m)^\intercal   d^{p}.
\end{align} 
The $(m,p)$th entry $\theta_{m p}$ in the weight matrix $\Theta \in R^{K \times K}$ is $0$ if  $d^m$ and $d^p$ are in the same dictionary and $1$ otherwise. The incoherence term promotes a discriminative dictionary such that $D_i$ and $D_j$ are as different as possible. The discriminative dictionaries are able to enhance the disaggregation performance.}

 {Given an aggregate test data $\hat{x}$, we aim to disaggregate the aggregate measurement into individual load $c$, denoted by $\hat{x}^c$. The objective function in the testing stage can be written as
\begin{align} \label{w} 
     \min_{w \in \mathbb{R}^q }  &  \lVert \hat{x} - \hat{D} \tilde{A} w \rVert_2       + \mu  \lVert w \rVert_1,  
     \end{align} 
where   we select a submatrix $\tilde{A}=[\tilde{A}_1;\cdots; \tilde{A}_n] \in \mathbb{R}^{K \times q}$ from $\hat{A}$. $\hat{D}$, $\hat{A}$ is the solution by solving the objective function \eqref{eqn:dictionary}. $\mu$ is a pre-defined hyper-parameter. The intuition is that some load  combinations are repetitive in the training data. We can select some representative combinations and disaggregate the aggregate measurement with respect to these combinations  to improve the disaggregation accuracy. Let $\hat{w}$ be the solution to \eqref{w}, then the estimated load  for load $c$ is $\hat{x}^c= \hat{D}_i \tilde{A}_i \hat{w}$. }

 \subsection{Bayesian Energy Disaggregation}
 
{ In \cite{YW21}, we  propose a   Bayesian method to deal with partial label data and provide the confidence measure of our disaggregation results. An overall framework is shown in Fig. \ref{fig3}. Given the training data $\bar{X}$ and   partial labels $\bm{Y}_\Omega$,  the proposed Bayesian method learns the posterior distribution of dictionaries and coefficients in the training stage. At the testing stage, the method learns the distributions of coefficients based  on the learned distributions of dictionaries. The distribution of $\hat{x}^c$ is then computed, where  $\hat{x}^c$ is the estimated power consumption of load $c$.  The mean of the distribution of $\hat{x}^c$ is used as the estimation of the load $c$ and the covariance is computed to measure the uncertainty.

The proposed method is based on a hierarchical probabilistic model. The prior distribution of the aggregate data ${x}_i$ can be written as 

 \begin{equation} \label{equation1}
\bar{x}_i = \sum_{c=1}^{C}{D}_{c} \bm{\omega}_i^c+\bm{\epsilon}_i 
\end{equation}
\begin{equation}\label{equation2}
\bm{\omega}_i^c = (\bm{z}_i^c \odot \bm{s}_i^c)y_i^c
\end{equation}
for all  $i=1,2,3,...,N$, $c=1,2,3,...,C$, 
where $\bm{\omega}_i^c \in \mathbb{R}^{K_c}$ is the coefficients for $D_c$, and $\bm{\epsilon}_i$ is the measurement noise. {In \eqref{equation2}, $\odot$ represents the element-wise product.} Let $\bm{d}^c_k$ denote the $k$th column in the dictionary ${D}_{c}$. $\bm{d}^c_k$ is sampled from a multivariate Gaussian distribution $\mathcal{N}(\bold{0},\frac{1}{\lambda_d} \bm{I}_P)$, where $\lambda_d$ is a pre-defined scalar, and $\bm{I}_P$ is an  identity matrix with size $P\times P$.  The  noise $\bm{\epsilon}_i$ is sampled from Gaussian $\mathcal{N}(\bold{0},\frac{1}{\gamma_\epsilon} \bm{I}_P)$.   One can see from (\ref{equation2}) that $\bm{\omega}_i^c$ is the element-wise product of $\bm{z}_i^c$ and $\bm{s}_i^c$ and then multiplied by $y_i^c$. $y_i^c$ is a binary variable sampled from a Bernoulli distribution and $y_i^c=1$ indicates that load $c$ exists in ${x}_i$. $\bm{z}_i^c$ is a binary vector. Let $\bm{z}_{ik}^c$  denote the $k$th entry of  $\bm{z}_i^c$. $\bm{z}_{ik}^c=1$  indicates $d_k^c$ is used to represent ${x}_i$ and 0 otherwise. $\bm{z}_{ik}^c$  is sampled from the Bernoulli distribution. Note that the Bayesian method is able to infer the actual dictionary size $K_c$ by gradually pruning the dictionary size based on  $\bm{z}_i^c$ in the training stage. Therefore, the Bayesian method is not sensitive to the selection of initial dictionary size. 
{$\bm{s}_i^c$ is sampled from   $\mathcal{N}(\bold{0},\frac{1}{\gamma_s^c} \bm{I}_{Kc})$.   We put Gamma priors on $\gamma_s^c$ and $\gamma_\epsilon$, respectively. { The Gamma priors are conjugate priors of the Gaussian distribution. If conjugate priors are selected, we can derive the analytical solution of the posterior distribution in the variational inference, which simplifies the updating process.}   Note that our model selects conjugate priors to simplify the updating process.} 

 Let $\bm{\Theta}$ denote all the latent variables. Given   $\bar{X}$ and partial labels $\bm{Y}_\Omega$, the objective is to obtain the posterior $P(\bm{\Theta}, \bm{Y}_{\bar{\Omega}} |\bm{X}, \bm{Y}_\Omega)$. From the Bayes theorem, 

\begin{equation}\label{eqn:bayes}
  P(\bm{\Theta}, \bm{Y}_{\bar{\Omega}} |\bar{X}, \bm{Y}_\Omega)=\frac{P(\bm{\Theta}, \bar{X}, \bm{Y})}{P(\bar{X},\bm{Y}_{\Omega})}
    \end{equation}
Because computing \eqref{eqn:bayes} directly is intractable, we use Gibbs sampling \cite{I12} to compute the posterior distribution. Gibbs sampling  sequentially samples from the conditional probability of one variable in $\bm{\Theta}$ and $\bm{Y}_{\bar{\Omega}}$ while keeping all other variables fixed. These conditional distributions have closed-form expressions because of the conjugate priors, which leads to an efficient updating process.

In the testing stage, given the aggregate test data $\hat{\bm{x}}$, the goal of our approach is to estimate  $\hat{{x}}^c$. A similar probabilistic model for  $\hat{\bm{x}}$ and $\hat{{x}}^c$ is described as:

\begin{equation} \label{xhat}
  \hat{{x}}  = \sum_{c=1}^{C}{D}_{c} (\bm{\hat{z}^c} \odot \bm{\hat{s}^c})\hat{y}^c+\hat{\bm{\epsilon}}\\
\end{equation}
 
\begin{equation}\label{eqn:hatxc}
 {\hat{{x}}^c} = {D}_c (\bm{\hat{z}^c} \odot \bm{\hat{s}^c})\hat{y}^c+\frac{\hat{\bm{\epsilon}}}{C}.
 \end{equation}
for all $c=1,...,C$, $k=1,...,K_c$.

The dictionary atom $d_k^c$ is sampled from learned distribution  $p({d_k^c}|\bm{X},\bm{Y}_\Omega)$ in the training stage. We also assume that $\hat{y}^c$ and $\hat{\bm{z}}^c$ are sampled from Bernoulli distributions. $\hat{\bm{s}}^c$ is sampled from $\mathcal{N}(\bold{0},\frac{1}{\gamma_s^c} \bm{I}_{K_c})$ and  $\hat{\bm{\epsilon}}$ is sampled from $\mathcal{N}(\bold{0},\frac{1}{\hat{\gamma}_{\epsilon}} \bm{I}_P)$. Gibbs sampling is also employed for computing probabilistic distributions of $\hat{y}^c$, $\hat{\bm{z}}^c$, $\hat{\bm{s}}^c$, and $\hat{\gamma}_\epsilon$.

{The per-iteration computational complexity of the Bayesian offline training is $\mathcal{O}(CK_cPN)$. The per-iteration computational complexity of the online testing   is $\mathcal{O}(CK_cP)$. Thus, the computational complexity scales linearly with respect to the number of loads. }

 \subsection{Uncertainty Modeling}
 Equipped with all learned posterior distributions, we then estimate the distribution of  $\hat{x}^c$. However, it is intractable to obtain the explicit expression for the distribution of  $\hat{x}^c$.   Monte-Carlo integration \cite{PBJ12} is employed to approximately compute the predictive mean and predictive variance.

Define
\begin{equation}
f (\bm{\Psi})  = {D}_c (\hat{\bm{z}}^c \odot \hat{\bm{s}}^c)\hat{y}^c
\end{equation}
where $\bm{\Psi}=\{{D}_c, \hat{\bm{z}}^c, \hat{\bm{s}}^c, {\hat{y}^c}, \hat{\gamma}_\epsilon\}$.
The predictive mean of ${\hat{{x}}^c}$ is computed by
\begin{equation}  \label{predictmean}
\begin{split}
E[{\hat{{x}}^c}]
                     &\approx\frac{1}{L}\sum_{l=1}^{l=L}  f (\bm{\Psi}^l) 
\end{split}
\end{equation}
where $L$ is the number of Monte-Carlo samples. {More Monte-Carlo samples increase the  estimation  accuracy, at the cost of higher computational burden. Our experiments show that $50$ Monte-Carlo samples suffice to provide accurate estimations of the predictive mean and the predictive variance.} $\bm{\Psi}^l$ is sampled  from the learned distributions of  variables in $\bm{\Psi}$.  $E[{\hat{{x}}^c}]$ is then used as the estimation of the power consumption of load C.   

The predictive covariance is approximated by

\begin{equation} \label{predictvariance}
\begin{split}
 \textrm{Var}[{\hat{{x}}^c}] 
 =&E[{\hat{x}^c}{\hat{x}^c}{}^T]-{E[{\hat{x}^c}]}{E[{\hat{x}^c}]}^T \\
 \approx &\frac{\bm{I}_P}{LC}\sum_{l=1}^{l=L} \frac{1}{\hat{\gamma}_{\epsilon}^{l}} +\frac{1}{L}\sum_{l=1}^{l=L}  f (\bm{\Psi}_l){ f (\bm{\Psi}^l)}^T  \\ -&(\frac{1}{L}\sum_{l=1}^{l=L}  f (\bm{\Psi}^l))(\frac{1}{L}\sum_{l=1}^{l=L}  {f (\bm{\Psi}^l)}^T) 
\end{split}
\end{equation}

\noindent Let  $\sigma_i$ ($i=1,..., P$)  denote all the singular values  of $\textrm{Var}[{\hat{{x}}^c}]$.
The uncertainty index $U_c$   for individual load $c$ and the uncertainty index $U_{\textrm{all}}$ for total estimated loads are computed  as
\begin{equation}\label{Uncertaintyindex1}
U_c = \Sigma_{i = 1}^P \sigma_i
\end{equation}
\begin{equation}\label{Uncertaintyindex2}
U_{\textrm{all}}  = \Sigma_{c = 1}^C U_c
\end{equation}
The intuition is that a large variance indicates higher uncertainty of the estimation. The uncertainty index is able to characterize the confidence level of disaggregation results. 

 }

\section{Numerical Experiment}
{ The performance of the proposed methods is evaluated on a partially labeled dataset. The dataset contains two industry loads and one solar generation. $N=360$ training samples and $M=300$ testing samples are generated. Even though the generated training samples contain up to three loads, each sample is annotated with only one label. The testing samples also contain up to three loads and have no label. In the following experiments, $\gamma$ represents the percentage of the training data that measure individual loads. For example, $\gamma=50\%$ denotes that $50\%$ training data labeled as load $c$  contain pure load $c$ and the remaining $50\%$ data contain other loads.   

}  

{  \subsubsection{Error Metrics}  Several metrics are employed to compute the disaggregation error.  The standard Root Mean Square Error (RMSE)   \cite{RQFKT17, PGWRA12} is defined as, 
\begin{align}
\text{RMSE}_c & =  \sqrt{\frac{\Sigma_{i=1}^{M}  \lVert \hat{x}^c_i -x^c_i \rVert_2^2}{P \times M}.  } 
\end{align}
where $\hat{x}^c_i, x^c_i  \in \mathbb{R}^P$ are the estimated   and the ground-truth load $c$ in  the $i$th testing sample, respectively. 
 
A new Total Error Rate (TER) is proposed to compute the disaggregation error of all the loads as follows, 
\begin{align}
\text{TER}   = &  \frac{  \Sigma_{i = 1}^{M} \Sigma_{c =1}^C  \min(   \lVert  \hat{x}^c_i - x_i^c \rVert_1 ,  \lVert  {x}^c_i \rVert_1 )  }{   \Sigma_{i= 1}^{M} \Sigma_{c = 1}^C \lVert x^c_i \rVert_1 }   
\end{align}

The Weighted Root Mean Square Error (WRMSE) is proposed to take the uncertainty index into account.  The weighted average disaggregation error is computed as,

\begin{equation}\label{eqn:WRMSE}
{\text{WRMSE}_c   =  \sqrt{\frac{\Sigma_{i=1}^{M}    \frac{\lVert \hat{x}^c_i -{x}^c_i \rVert_2^2}{U_c(\hat{x}^c_i)} }{P\Sigma_{i=1}^{M} \frac{1}{U_c(\hat{x}^c_i)} }  } }
\end{equation}

\noindent where   $U_c(\hat{x}^c_i)$ denotes the uncertainty index of $\hat{x}^c_i$.   A larger  $U_c(\hat{x}^c_i)$  represents a less reliable estimation. If the estimated loads with higher disaggregation errors are accompanied by larger uncertainty indices, the RMSE$_c$ could be much larger than WRMSE$_c$. The scenario that RMSE$_c$ is much larger than WRMSE$_c$ indicates that the unreliable estimation results are correctly flagged by higher uncertainty indices.

\subsubsection{Methods}  Our deterministic EDS method in \cite{LYWW20} is abbreviated as ``D-EDS.'' Our Bayesian EDS method in \cite{YW21} is abbreviated as ``B-EDS.''  Two other existing methods are employed for comparison. The work in \cite{KBNY10} that is based on discriminative sparse coding is abbreviated as ``DDSC,'' and the work \cite{RQFKT17} based on sum-to-k matrix factorization  is abbreviated as ``sum-to-k''. Because we set the Monte-Carlo samples $L=50$ in our method B-EDS, then D-EDS, DDSC and sum-to-k are averaged over 50 runs for 
a fair comparison.  The comparisons  of disaggregation performance of B-EDS, D-EDS, DDSC and sum-to-k are shown in Table \ref{table1}. $\gamma=70\%$. Note that all the existing works such as DDSC and sum-to-k methods require fully labeled data to obtain accurate estimation. 
Directly applying the existing methods 
to partially labeled data leads to a low disaggregation accuracy.  The proposed two approaches B-EDS and D-EDS are designed for  partially labeled data and can achieve state-of-the-art disaggregation performance. 
Between these two methods, the disaggregation accuracy of B-EDS is slightly better. Moreover, one can see from Table I that the WRMSE$_c$ is much smaller than the corresponding RMSE$_c$. As we discussed above, this means that those estimations with larger disaggregation errors also have large uncertainty indices. This validates the effectiveness of applying the proposed uncertainty index to measure the reliability of the disaggregation results.

The major advantage of B-EDS over D-EDS is that B-EDS is able to measure the confidence level of disaggregation results from the uncertainty index. We provide five case studies to verify the performance of uncertainty modeling   of B-EDS.

\begin{itemize}
\item Case 1: we select  test data from the testing datasets in Table I and this test data contains three types of loads.

\item Case 2:  the test data is as same as the data in Case 1 with an additional Gaussian noise $\mathcal{N}(0,4^2)$ in each entry. 

\item Case 3: the test data is as same as the data in Case 1 with an additional Gaussian noise $\mathcal{N}(0,6^2)$ in each entry. 

\item Case 4: the test data only contains one solar generation, but the pattern of solar generation is different from the solar patterns in the training data.

\item Case 5: the test data contains the same load 1 and 2 as those in Case 1,  and as well as a solar generation with a pattern different from the solar patterns in the training data. 
\end{itemize}

Figs.~\ref{fig_4} and ~\ref{fig_5} show the disaggregation performance of D-EDS on Case1 and Case 4. 
The aggregate measurement  is shown in   (a), and the disaggregation results are shown in (b)-(d) in both figures. In Case 1, the disaggregation results by D-EDS follow the actual load pattern. In Case 4, the disaggregation result of the solar generation does not follow the actual solar pattern because it is different from the learned pattern from the training data. In both cases, the disaggregation results contain some errors. That motivates using the Bayesian approach to compute a probabilistic  distribution of load consumption rather than computing one deterministic estimation. 


{Fig.~\ref{fig_uncertainty} shows the 
  disaggregation performance of B-EDS on these five cases, and Table II shows the corresponding uncertainty index. 
  Each subfigure in Fig.~\ref{fig_uncertainty} plots the ground-truth load, the estimated load and the $99.7\%$ confidence interval of the estimated load. One can see 
  that in Cases 1-3, although there are some errors in the disaggregation results, the ground-truth loads are within the confidence interval. Moreover, the estimation errors increase slightly when the noise levels increase.  Correspondingly, Table II shows that the total uncertainty indices in Case 1-3 also increase as the noise level increases, which indicates the effectiveness of using the uncertainty index to characterize  the uncertainty in the estimation. In Case 4 and Case 5, because the patterns of solar generation are far from the patterns in the training data, the ground-truth load consumption may not fall inside the confidence interval (especially Case 5). 
  The uncertainty indices in Table II also increase significantly, indicating that the estimated results are less reliable in these cases. 
  Therefore, 
  the users can use the uncertainty index to evaluate the accuracy of the disaggregation results. 
Table II also compares the TER of B-EDS and D-EDS, and B-EDS has  a smaller disaggregation error of D-EDS.}

\begin{figure*}[!ht]  
     \centering
     \subfigure[]	{\includegraphics[width=0.24\textwidth]{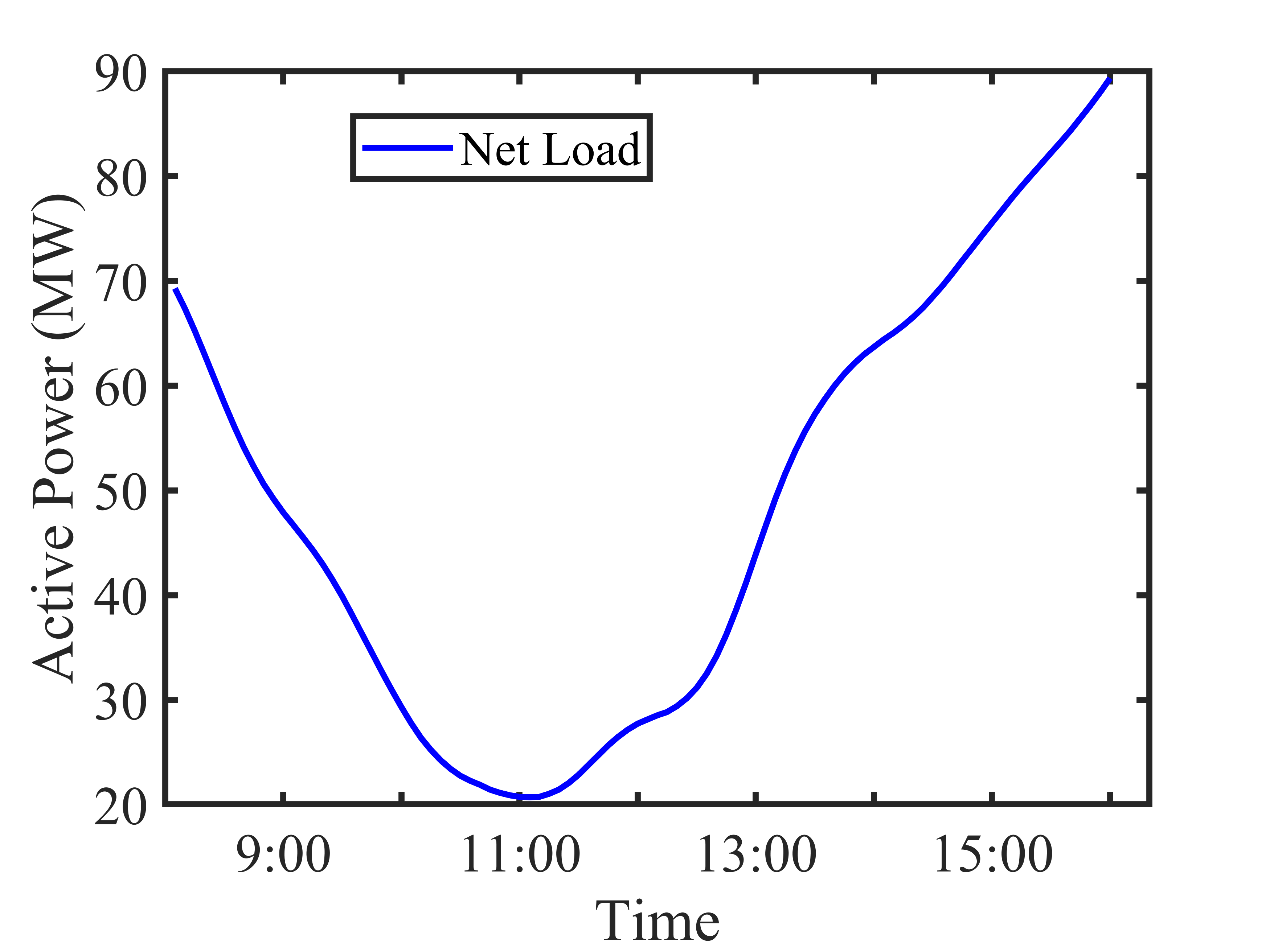}}
    \subfigure[]{	\includegraphics[width=0.24\textwidth]{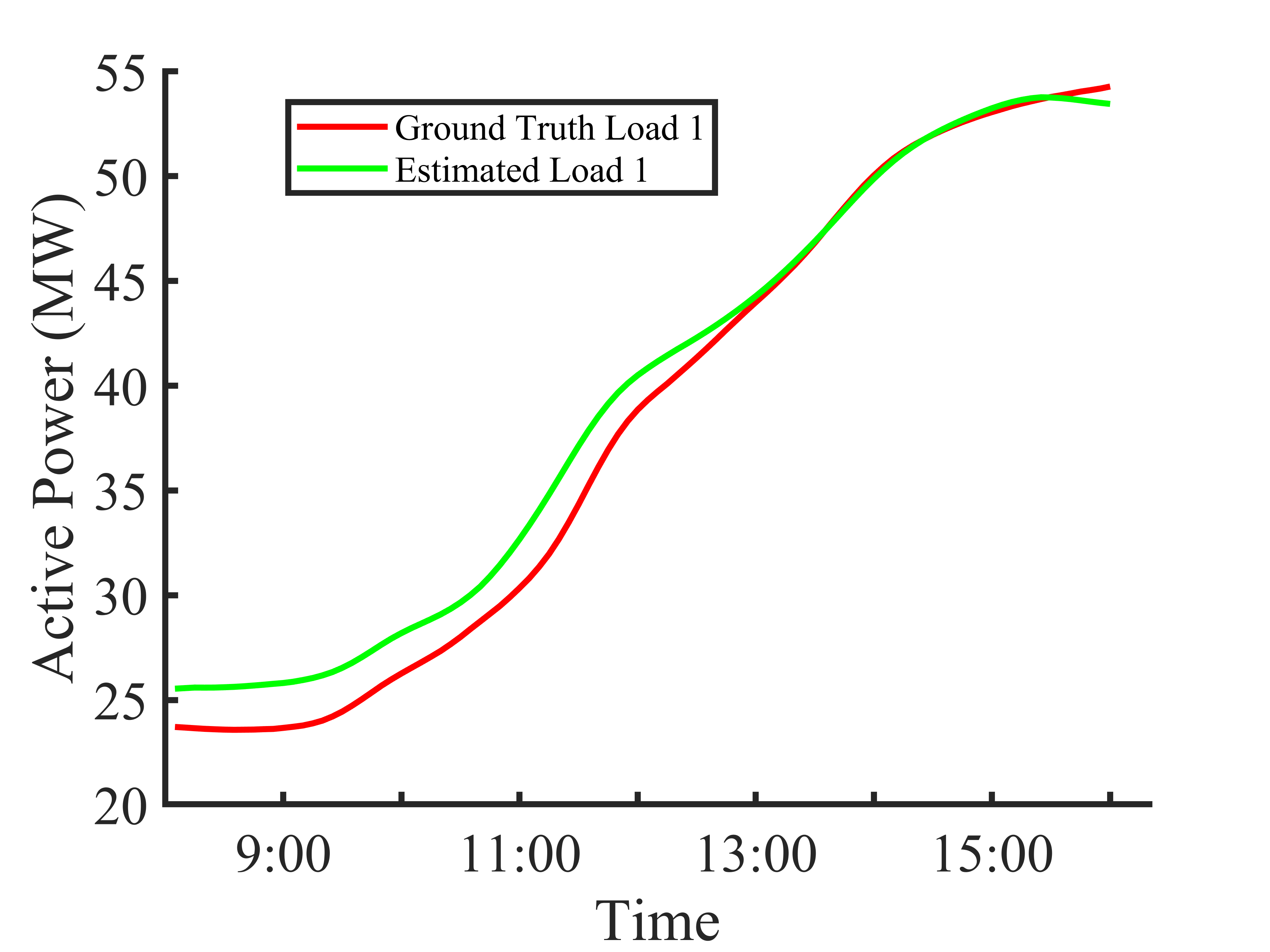}} 
    \subfigure[]{	\includegraphics[width=0.24\textwidth]{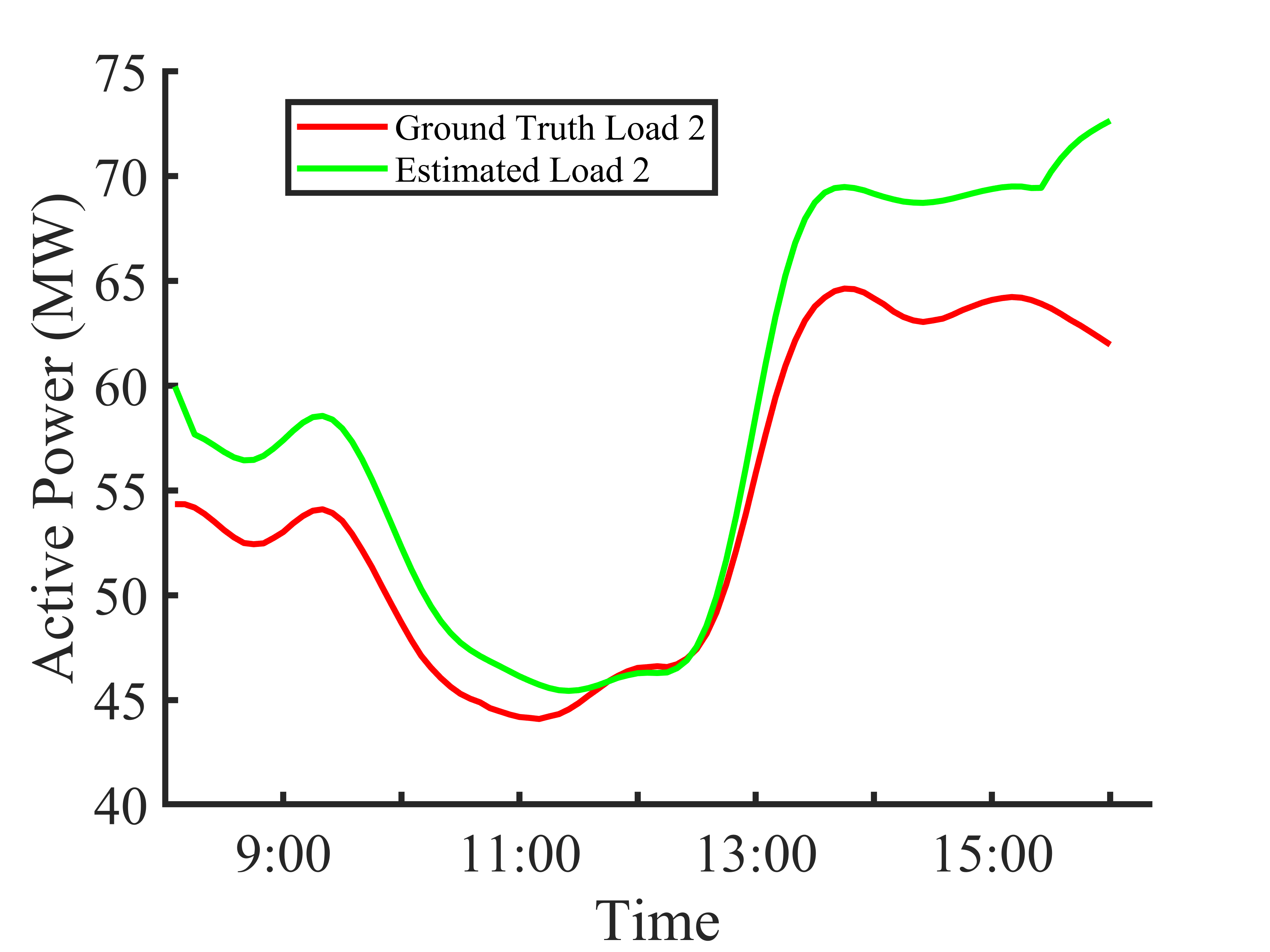}}
\subfigure[]{	\includegraphics[width=0.24\textwidth]{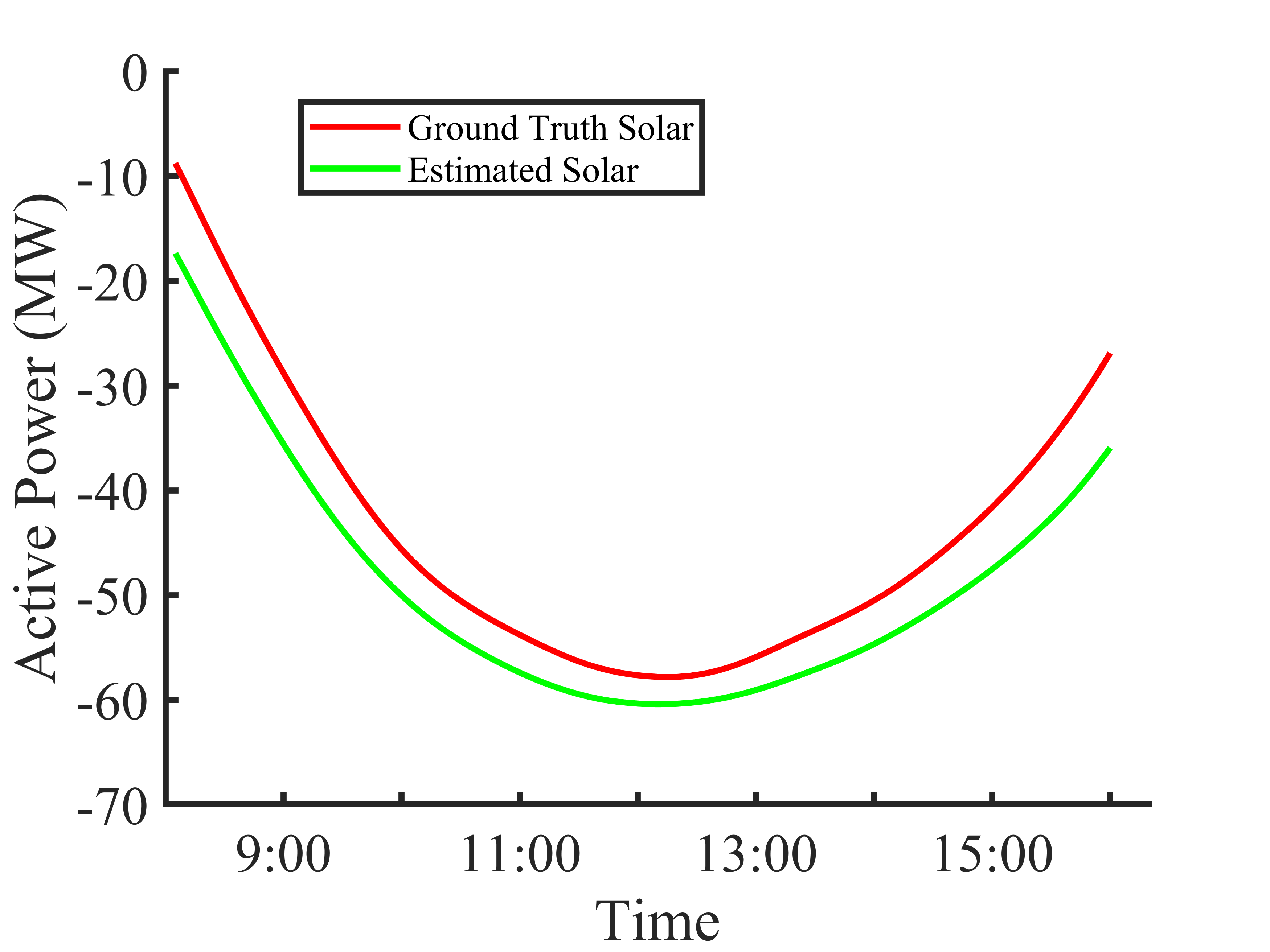}} 
    	\caption{ {Disaggregation performance of D-EDS on Case 1.(a) Net load. (b) The ground-truth  and disaggregated load 1.(c) The ground-truth   and disaggregated load 2. (d) The ground-truth   and disaggregated solar.} } \label{fig_4}
    \end{figure*}
    
\begin{figure*}[!ht]  
     \centering
     \subfigure[]	{\includegraphics[width=0.24\textwidth]{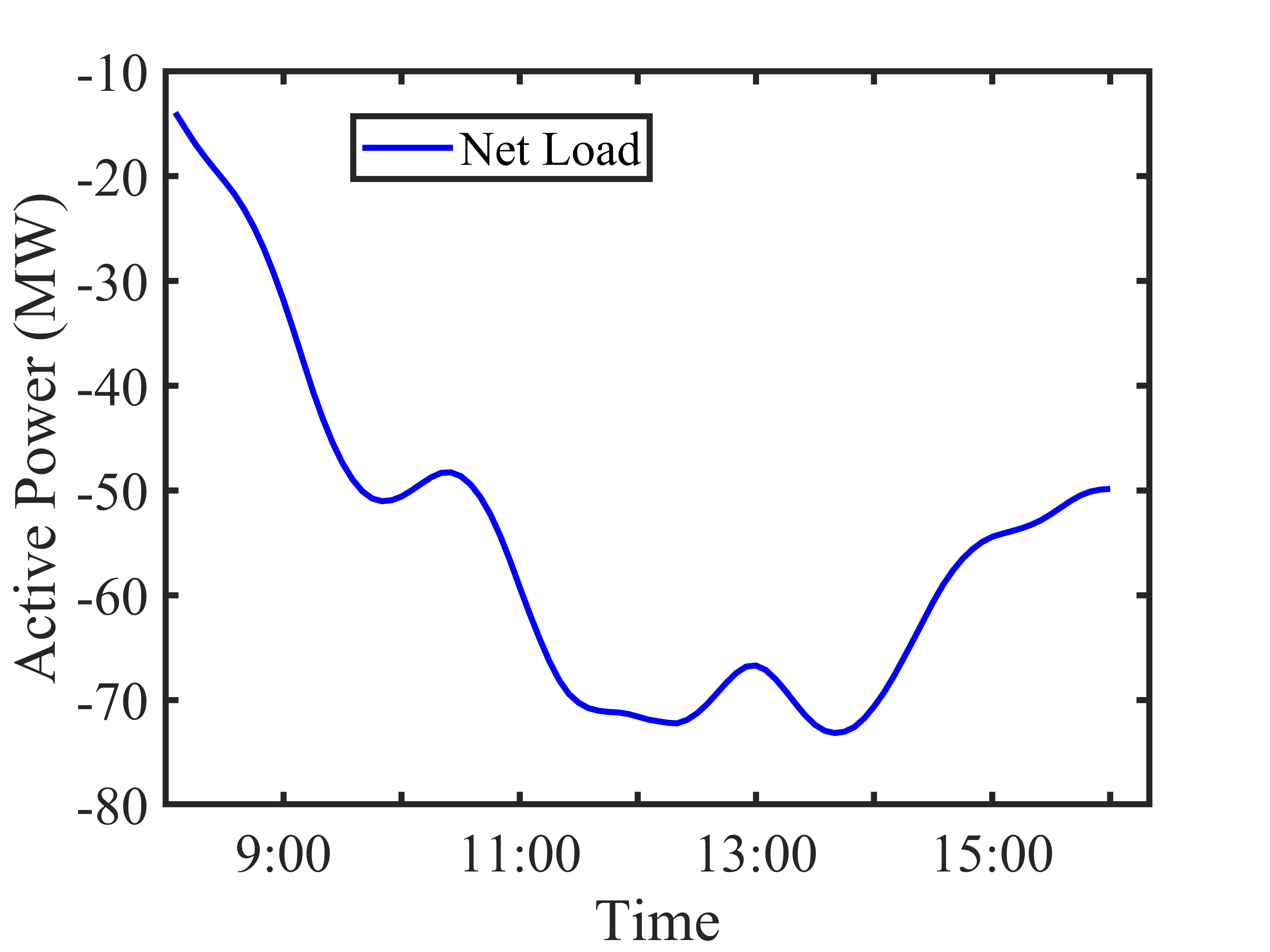}}
    \subfigure[]{	\includegraphics[width=0.24\textwidth]{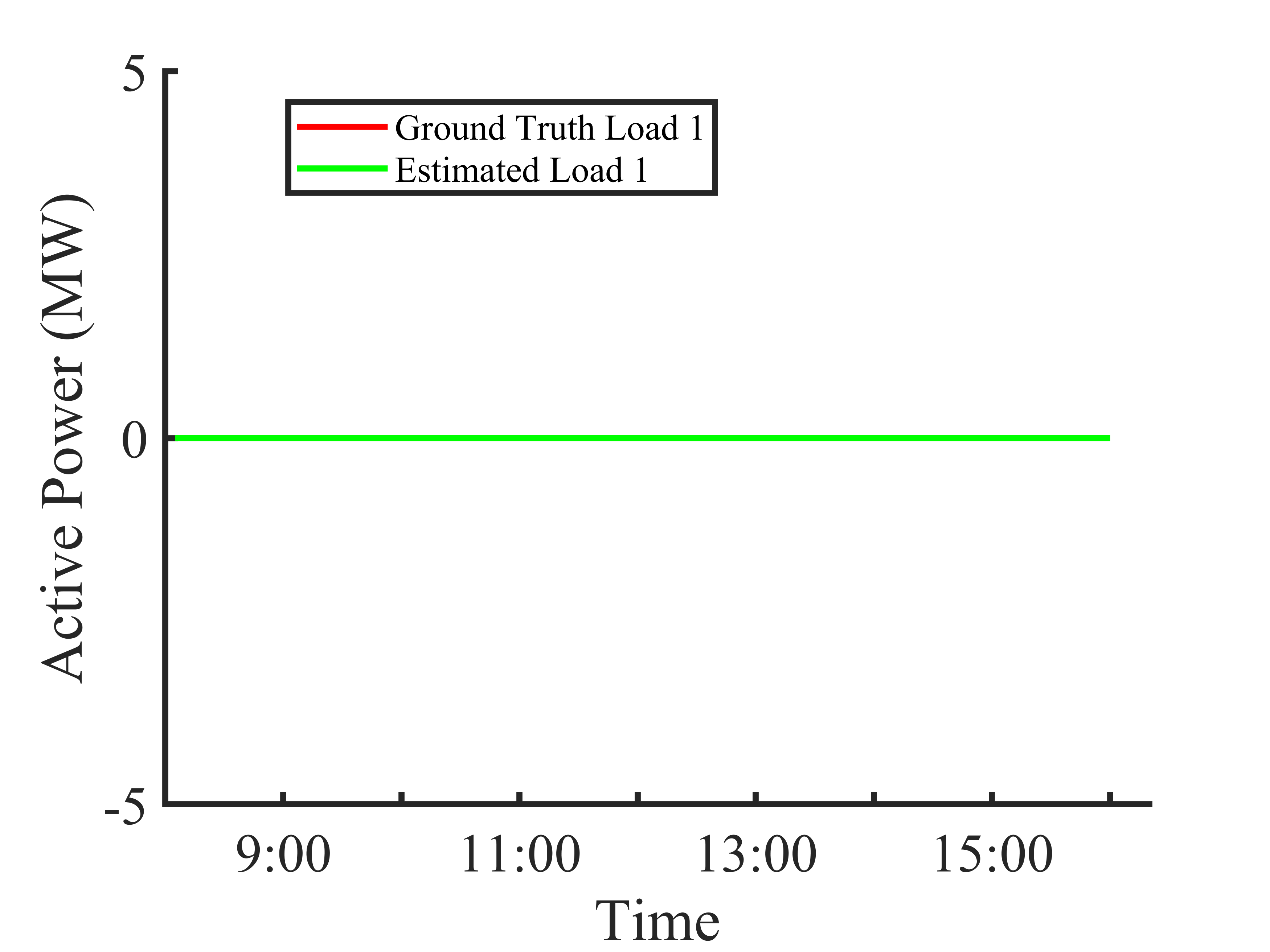}} 
    \subfigure[]{	\includegraphics[width=0.24\textwidth]{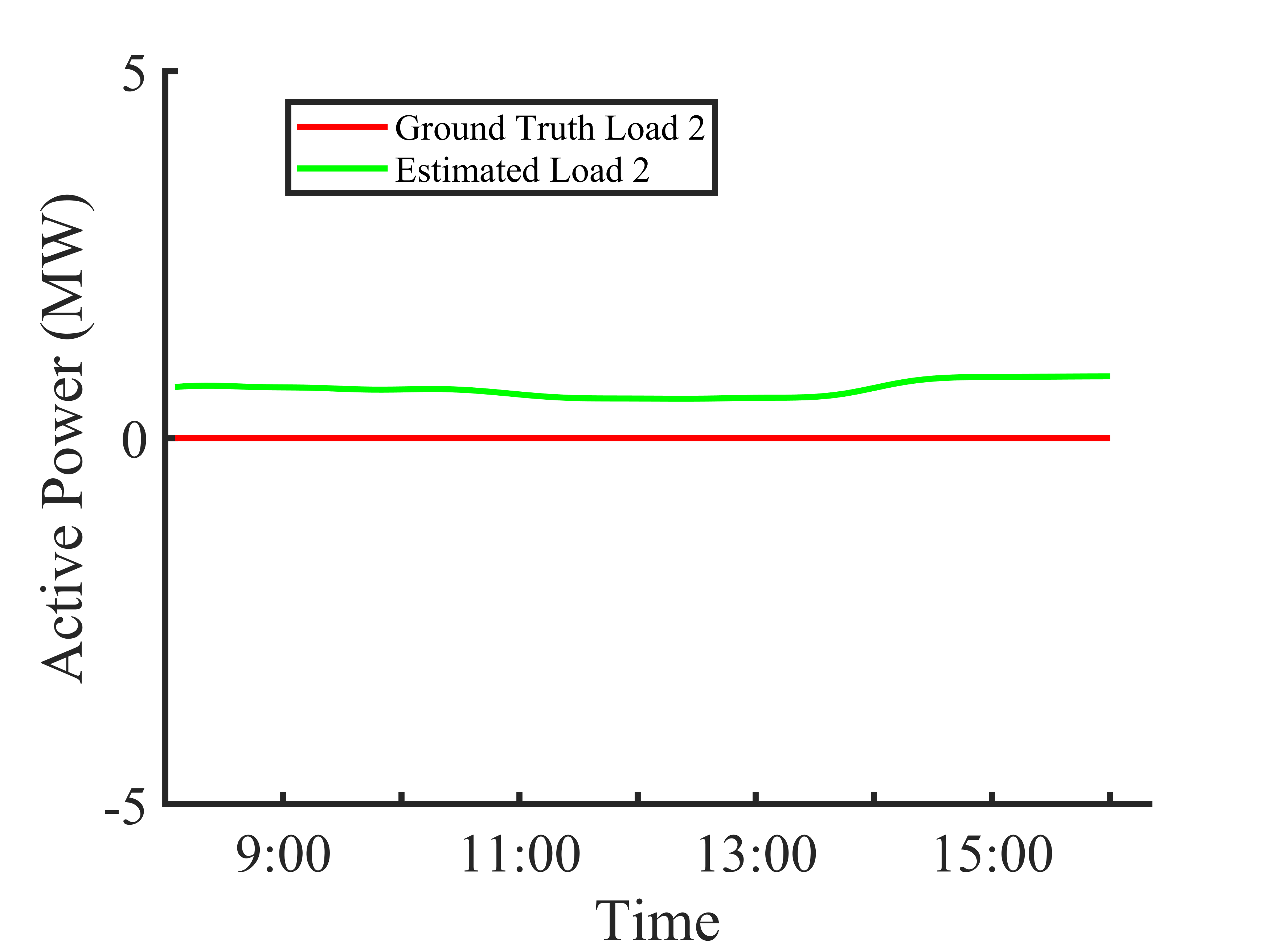}}
\subfigure[]{	\includegraphics[width=0.24\textwidth]{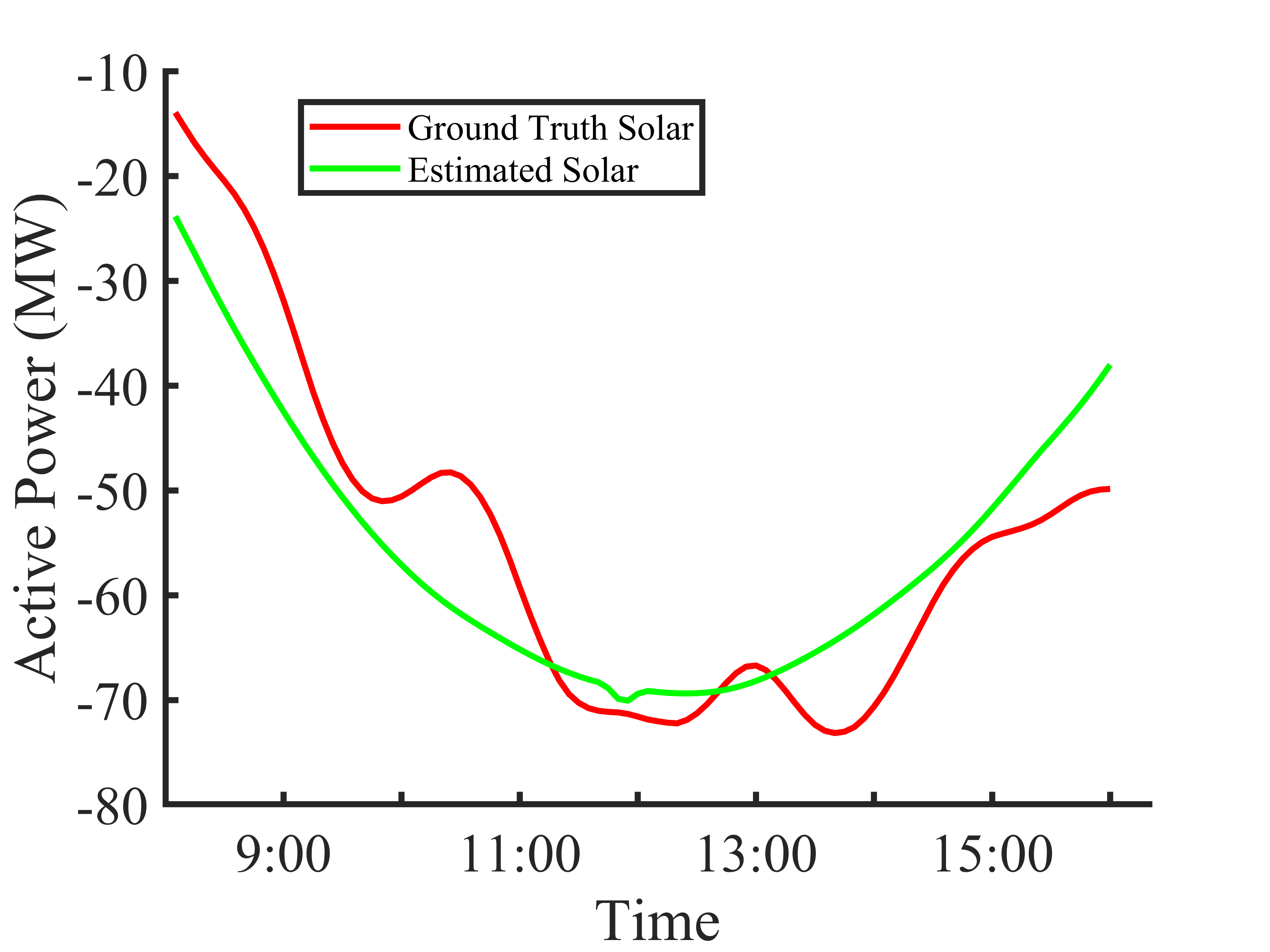}} 
    	\caption{ {Disaggregation performance of D-EDS on Case 4.(a) Net load. (b) The ground-truth   and disaggregated load 1.(c) The ground-truth   and disaggregated load 2. (d) The ground-truth  and disaggregated solar.} } \label{fig_5}
    \end{figure*}

    \begin{table}[]
\centering
\caption{Comparison between B-EDS, D-EDS, sum-to-k and DDSC methods on disaggregation accuracy}
\label{table1}
\begin{tabular}{lllll}
 \hline          & B-EDS    & D-EDS  & sum-to-k  & DDSC    \\
 \hline RMSE$_1$  & 6.20 & 6.62  & 13.17 & 22.77   \\
 \hline RMSE$_2$  & 5.19  & 6.34  & 11.35 & 23.86   \\
 \hline RMSE$_3$  & 5.82 & 4.65 & 10.70 & 13.49   \\
 \hline WRMSE$_1$ & 0.16 &  -      &  -       &     -    \\
 \hline WRMSE$_2$ & 0.13 &   -     &  -       &   -      \\
 \hline WRMSE$_3$ & 0.13 &    -    &  -       &   -      \\
 \hline TER      & 8.97\%   & 9.95\% & 20.61\% & 37.12\% \\
  \hline 
\end{tabular}
\end{table}

} 

\begin{figure*} [h!]
    \centering
    \subfigure[]{\includegraphics[width=0.18\textwidth]{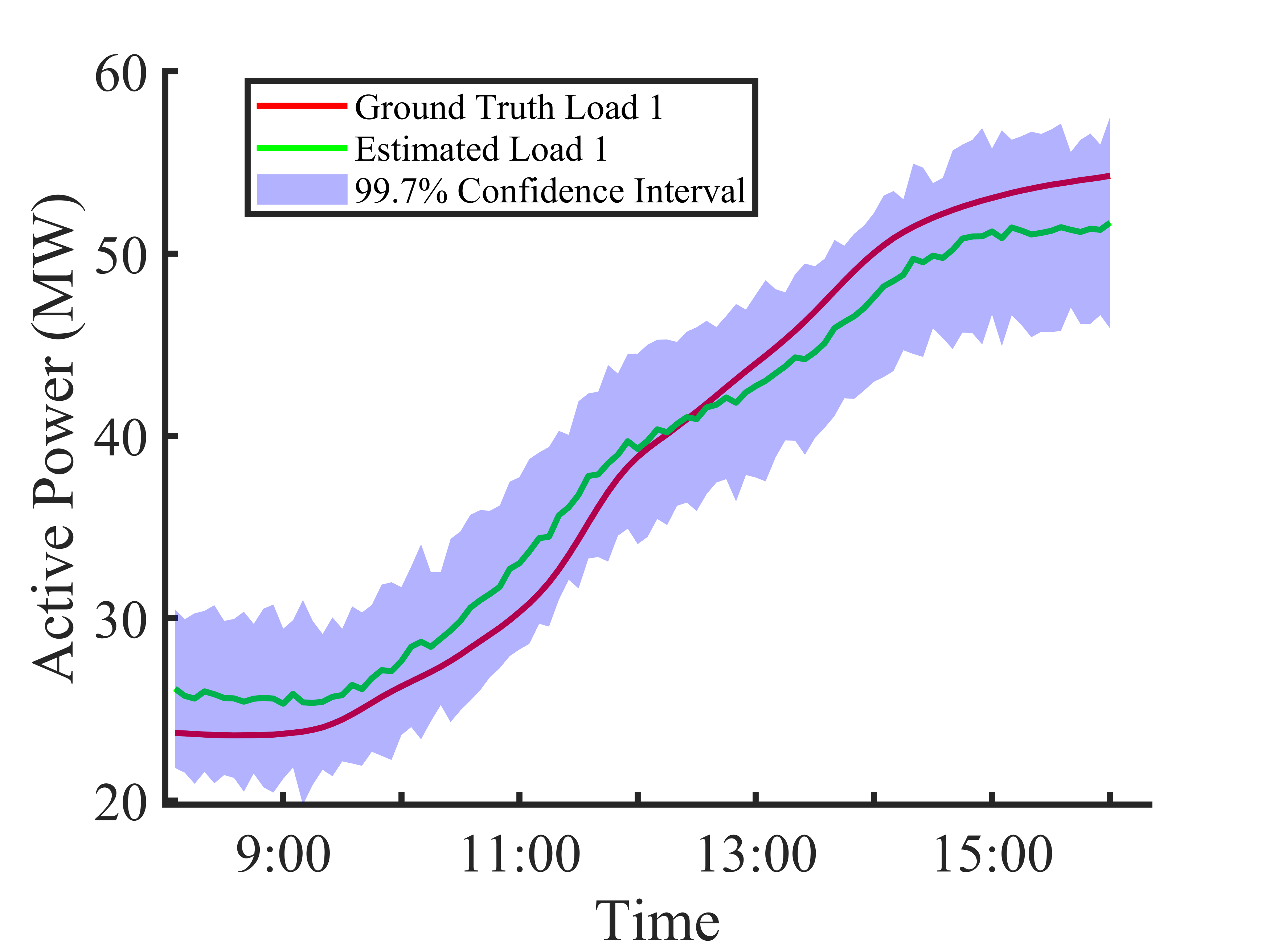}} 
    \addtocounter{subfigure}{+2}
   \subfigure[]{\includegraphics[width=0.18\textwidth]{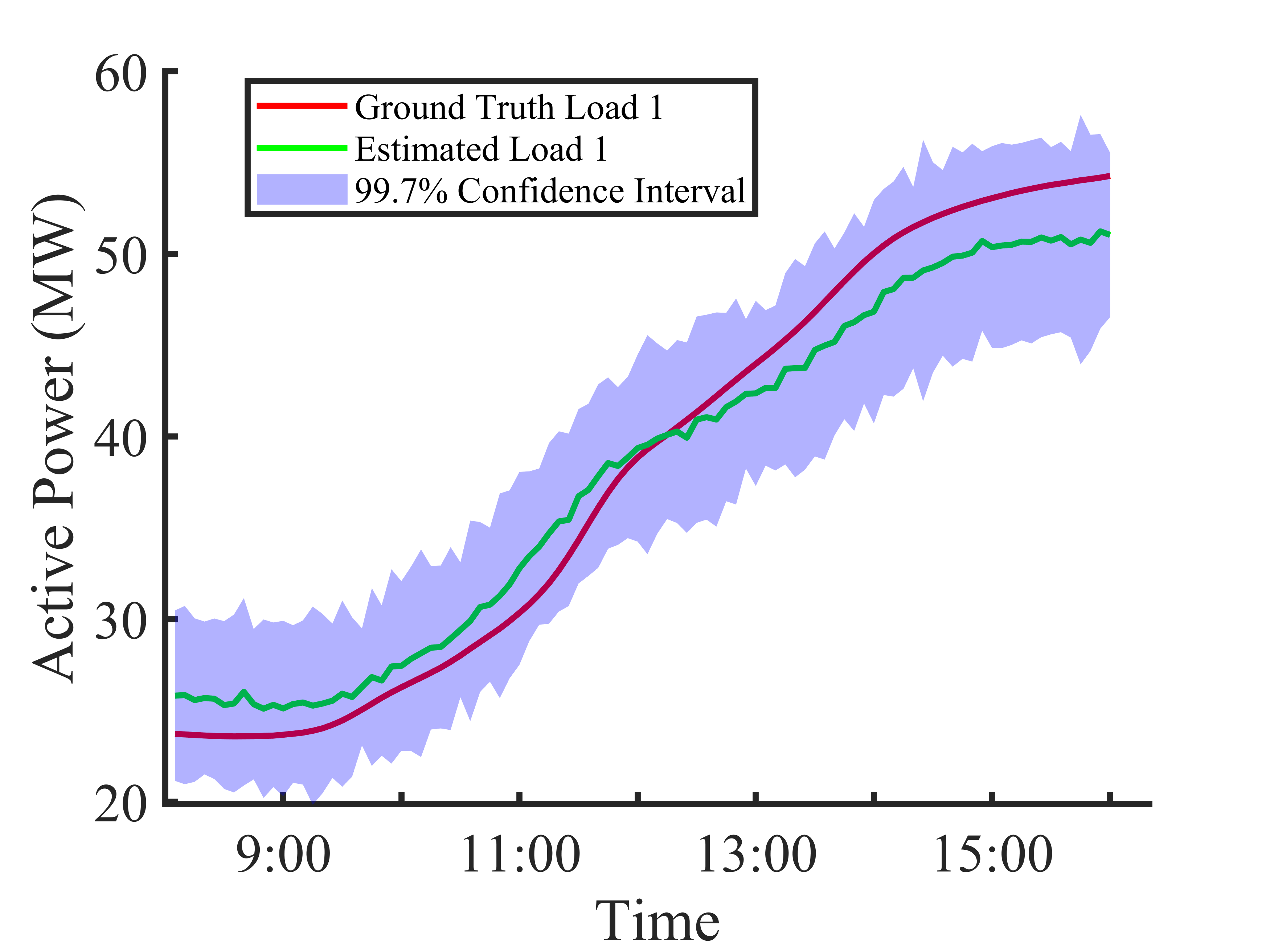}} 
    \addtocounter{subfigure}{+2}
    \subfigure[]{\includegraphics[width=0.18\textwidth]{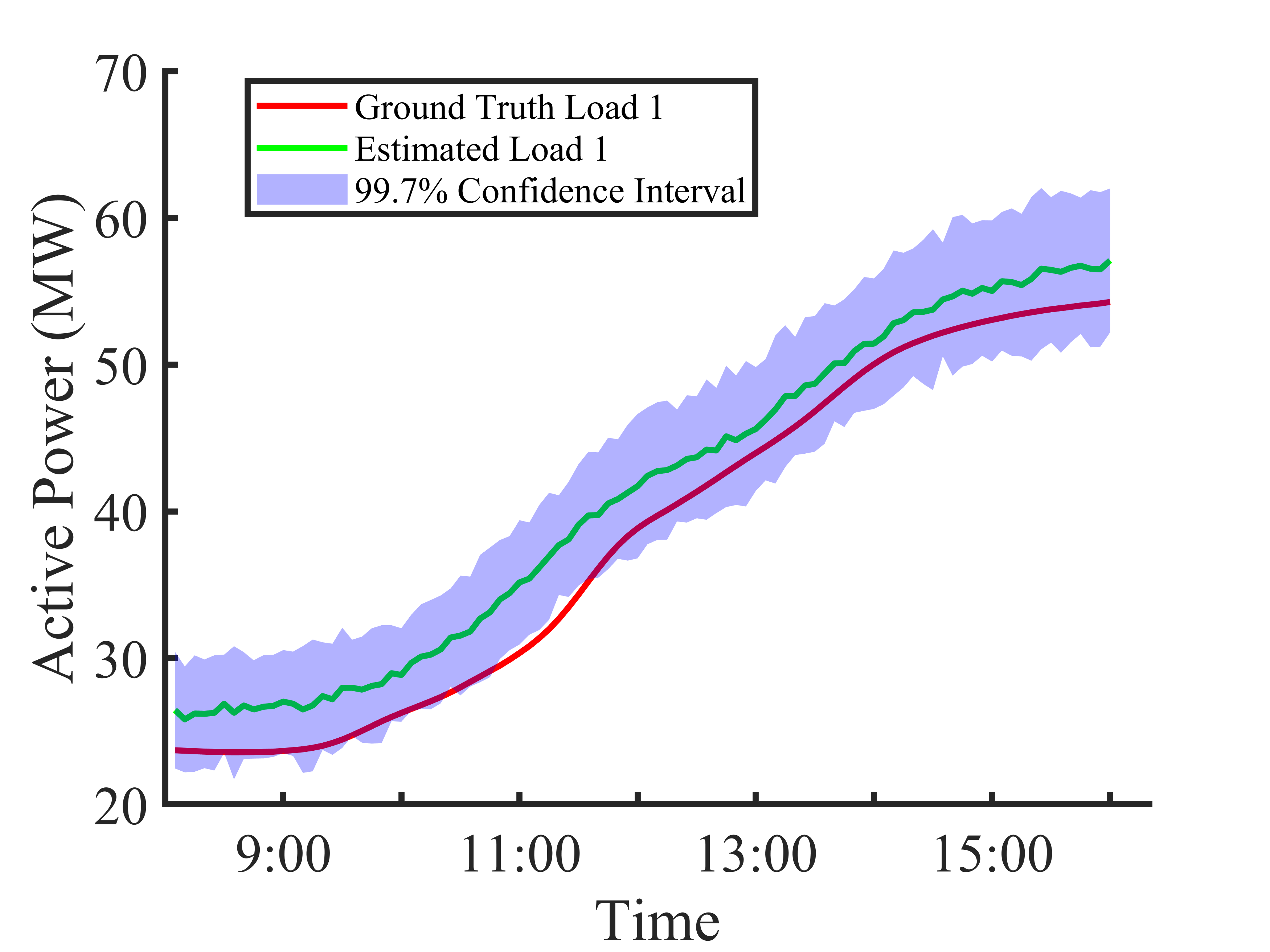}}
    \addtocounter{subfigure}{+2}
    \subfigure[]{\includegraphics[width=0.18\textwidth]{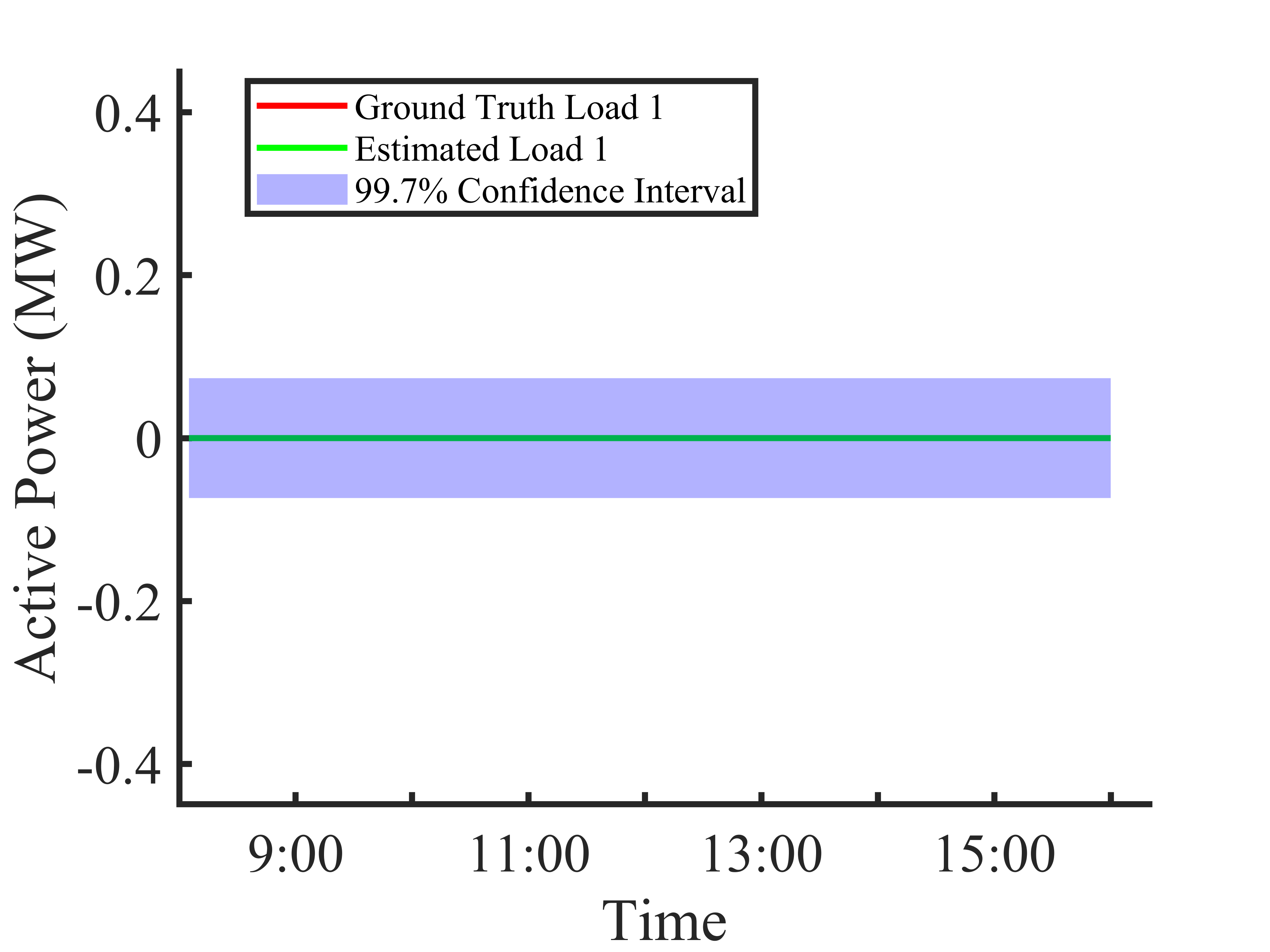}}
    \addtocounter{subfigure}{+2}
     \subfigure[]{\includegraphics[width=0.184\textwidth]{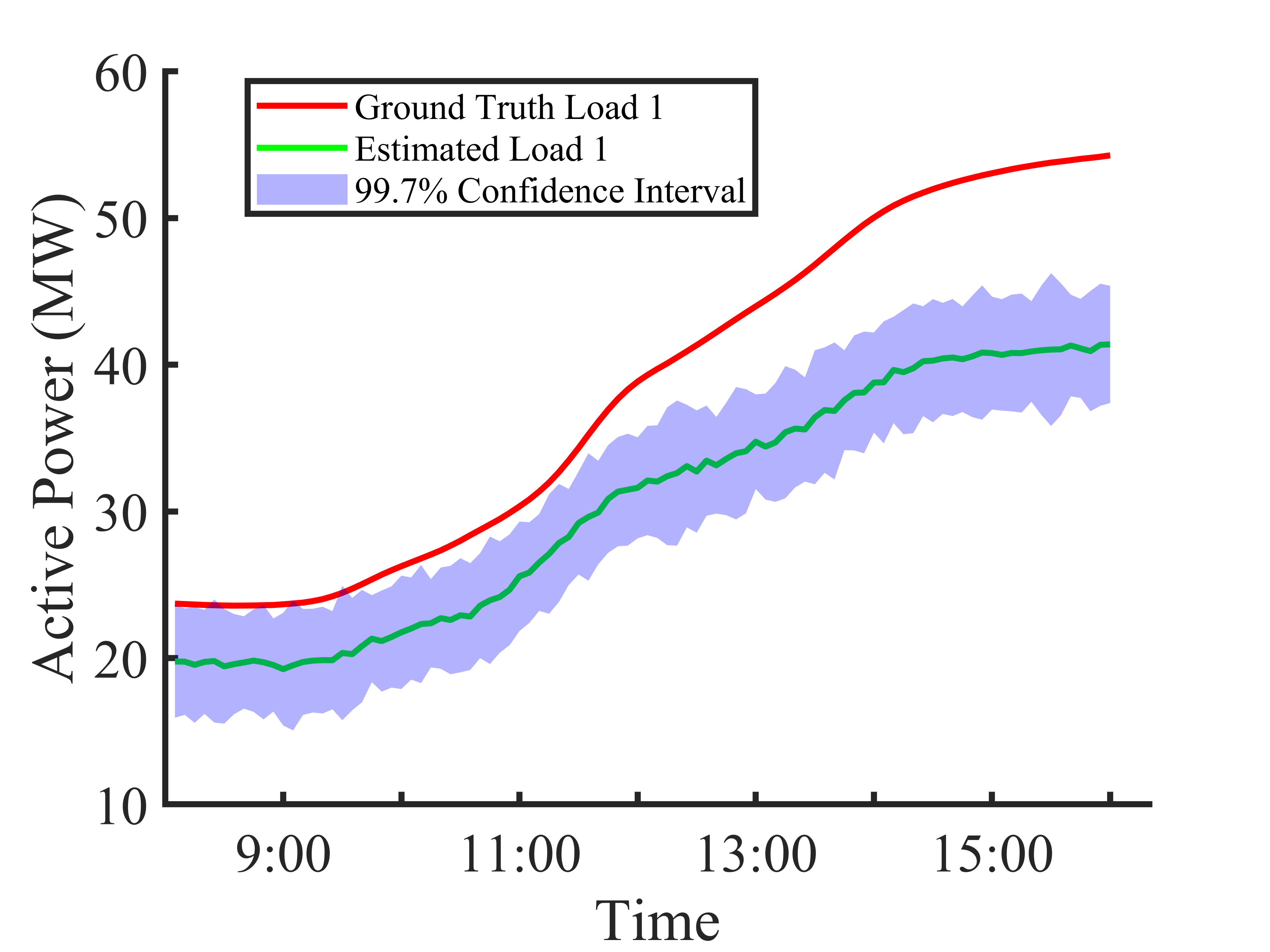}}

     \addtocounter{subfigure}{-12}
     \subfigure[]{\includegraphics[width=0.18\textwidth]{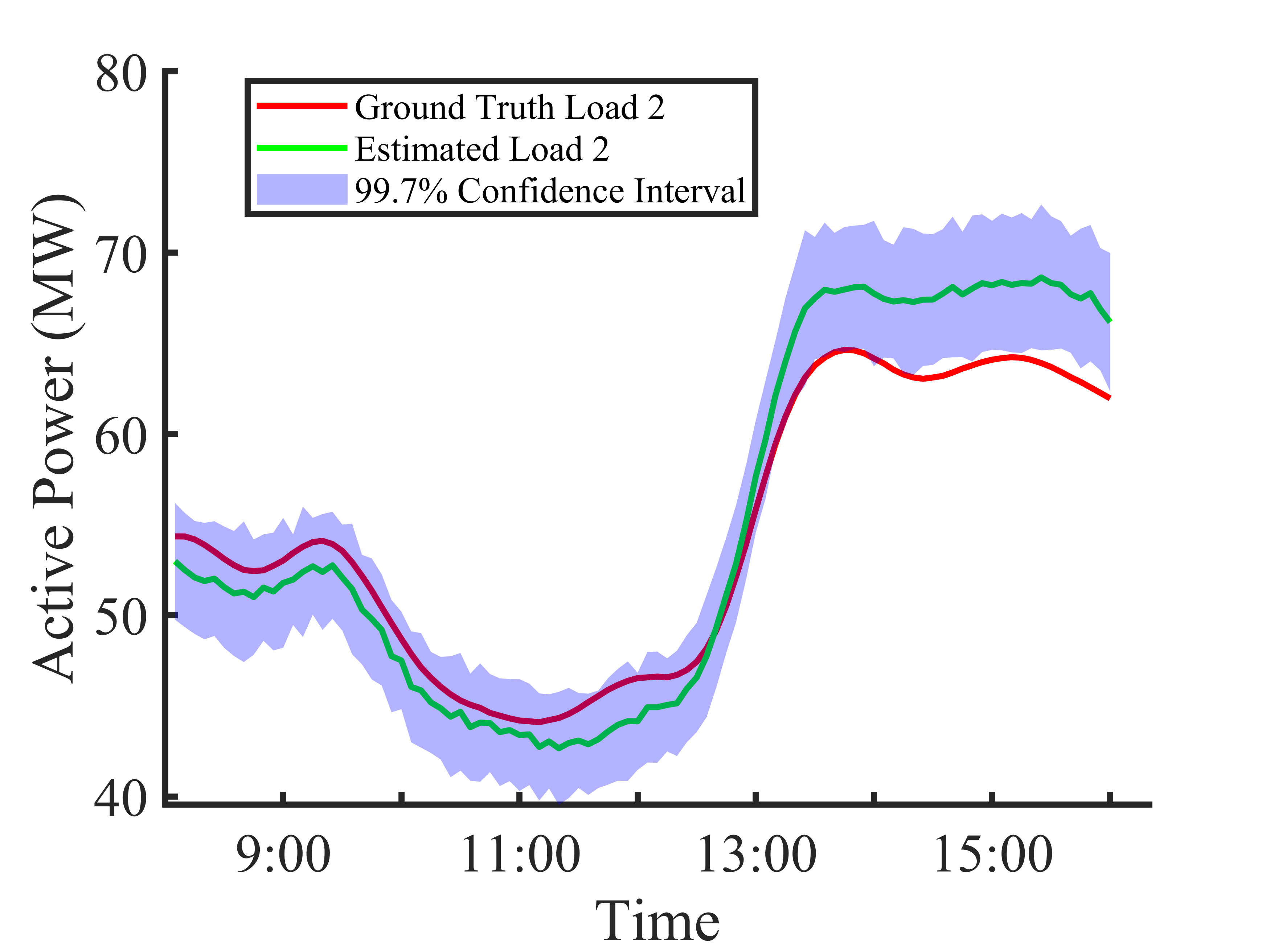}} 
      \addtocounter{subfigure}{+2}
    \subfigure[]{\includegraphics[width=0.18\textwidth]{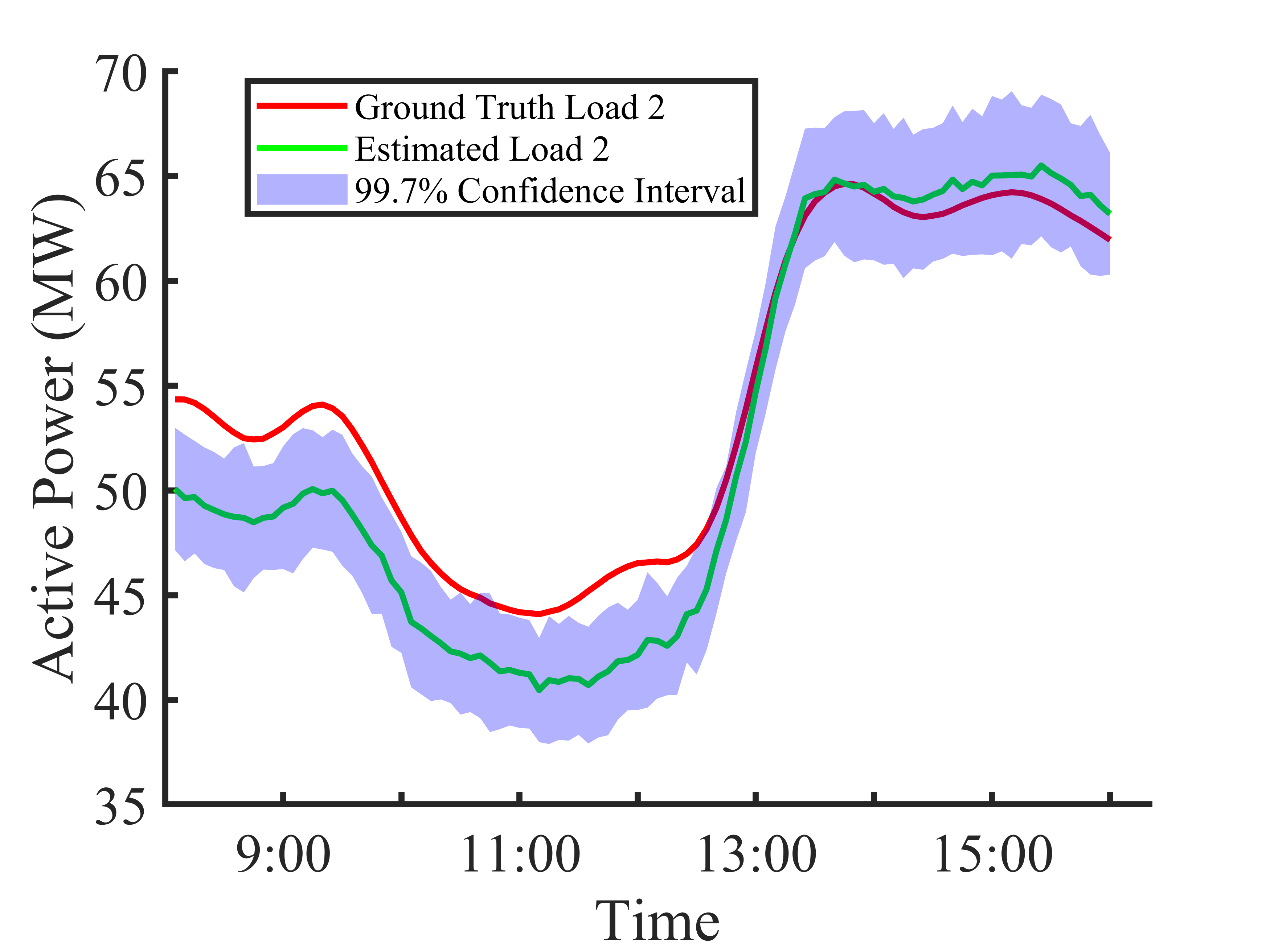}} 
          \addtocounter{subfigure}{+2}
    \subfigure[]{\includegraphics[width=0.18\textwidth]{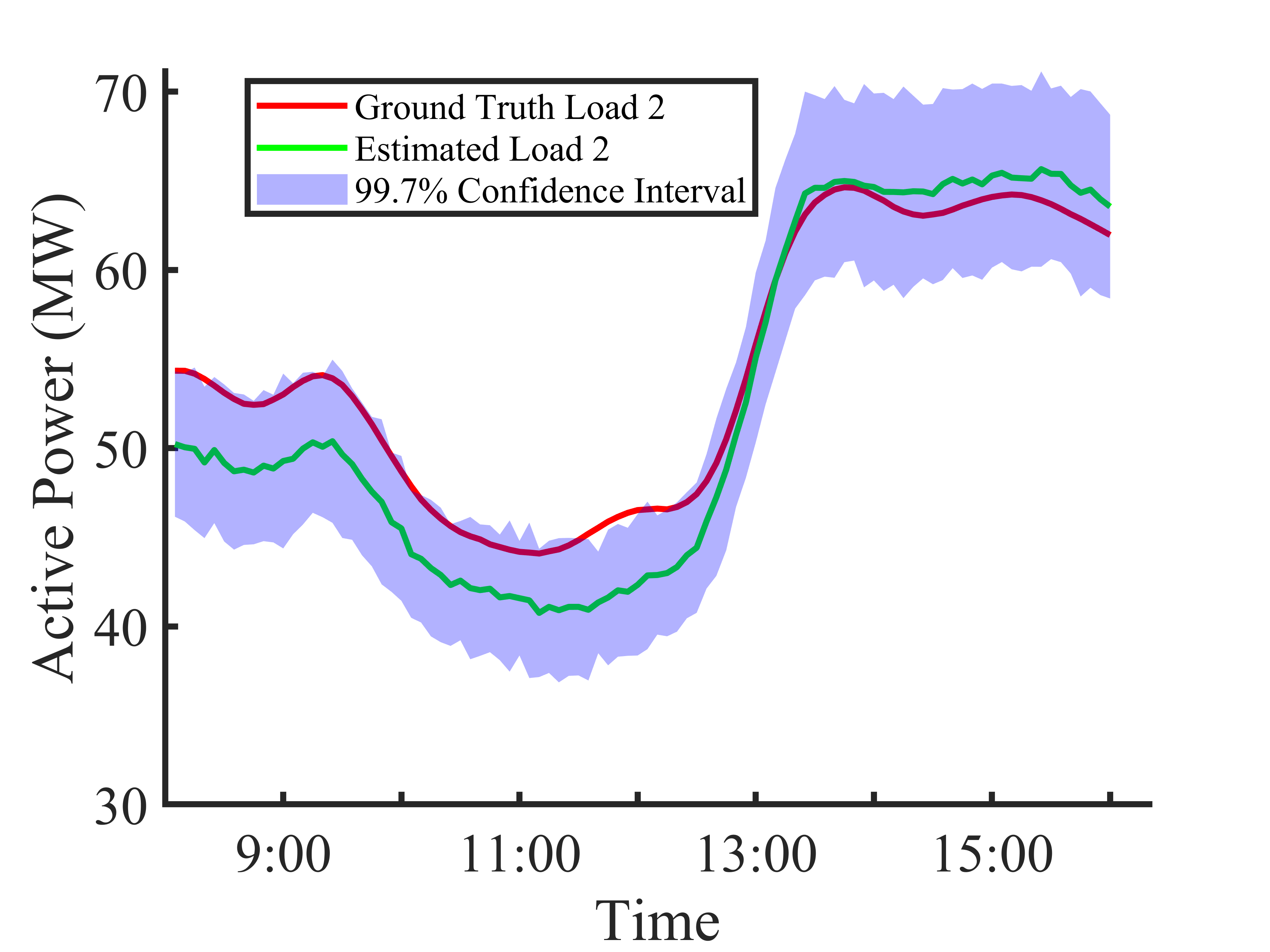}}
          \addtocounter{subfigure}{+2}
    \subfigure[]{\includegraphics[width=0.18\textwidth]{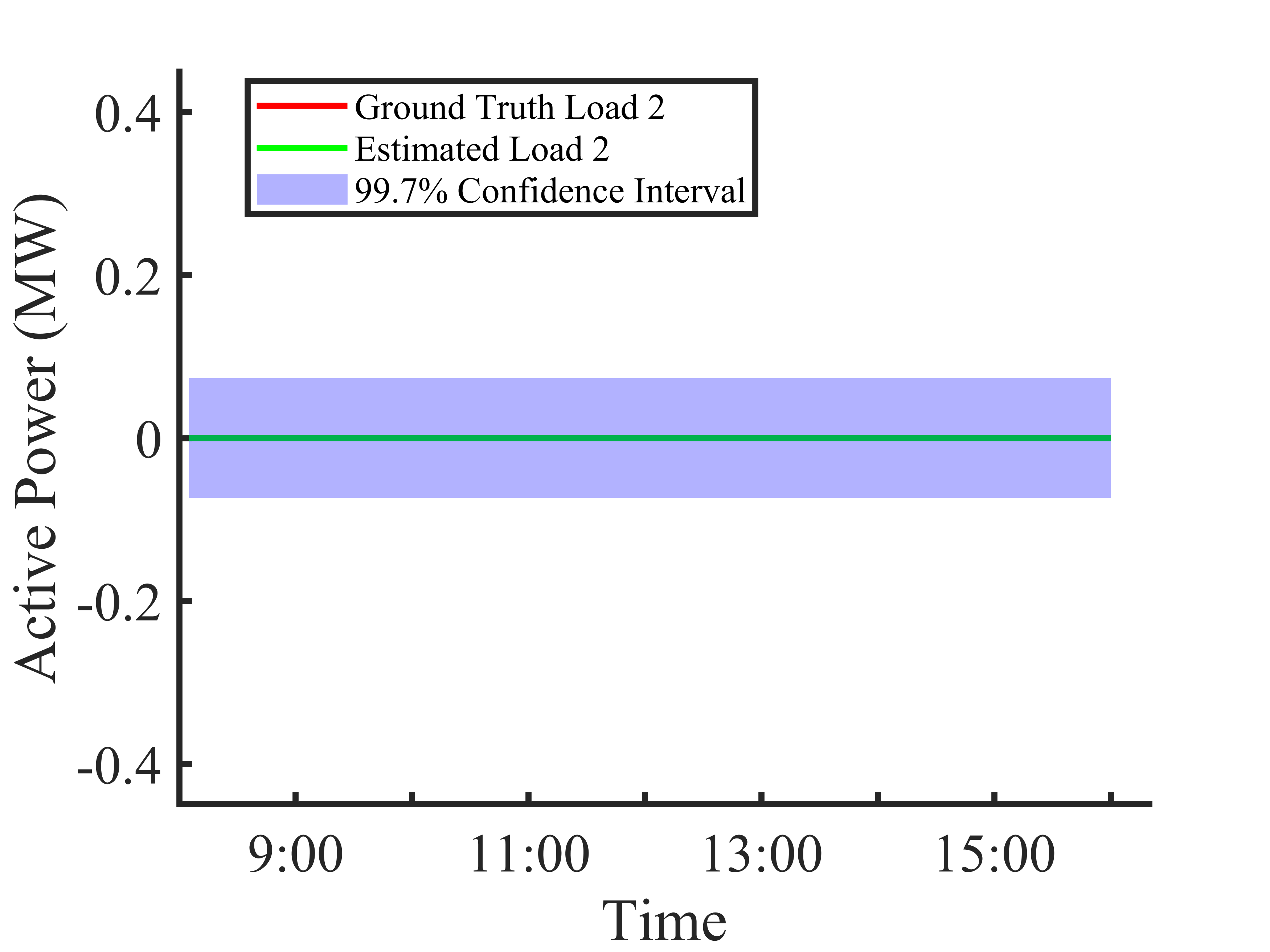}}
          \addtocounter{subfigure}{+2}
     \subfigure[]{\includegraphics[width=0.184\textwidth]{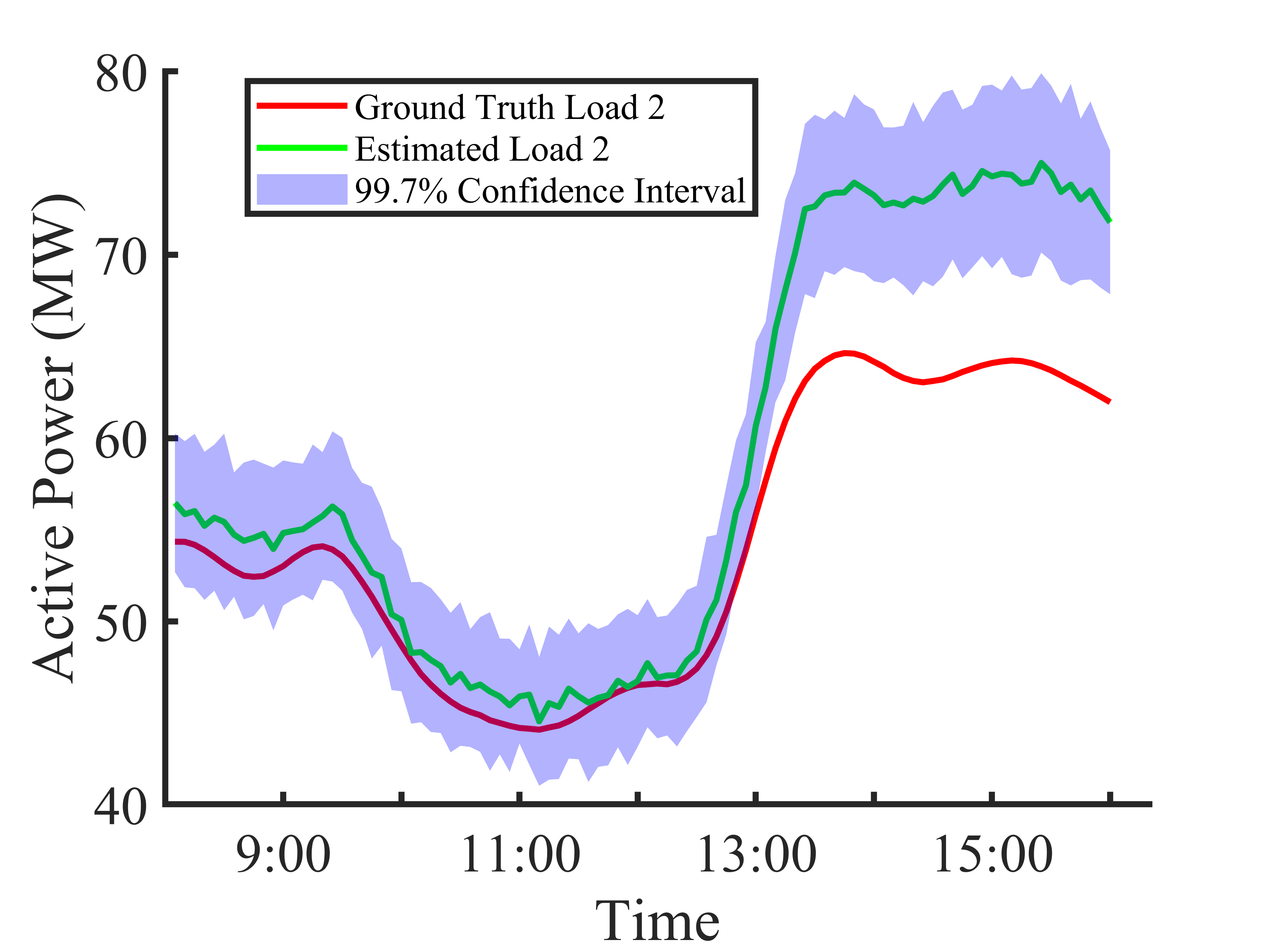}}
         \addtocounter{subfigure}{-12}
     \subfigure[]{\includegraphics[width=0.18\textwidth]{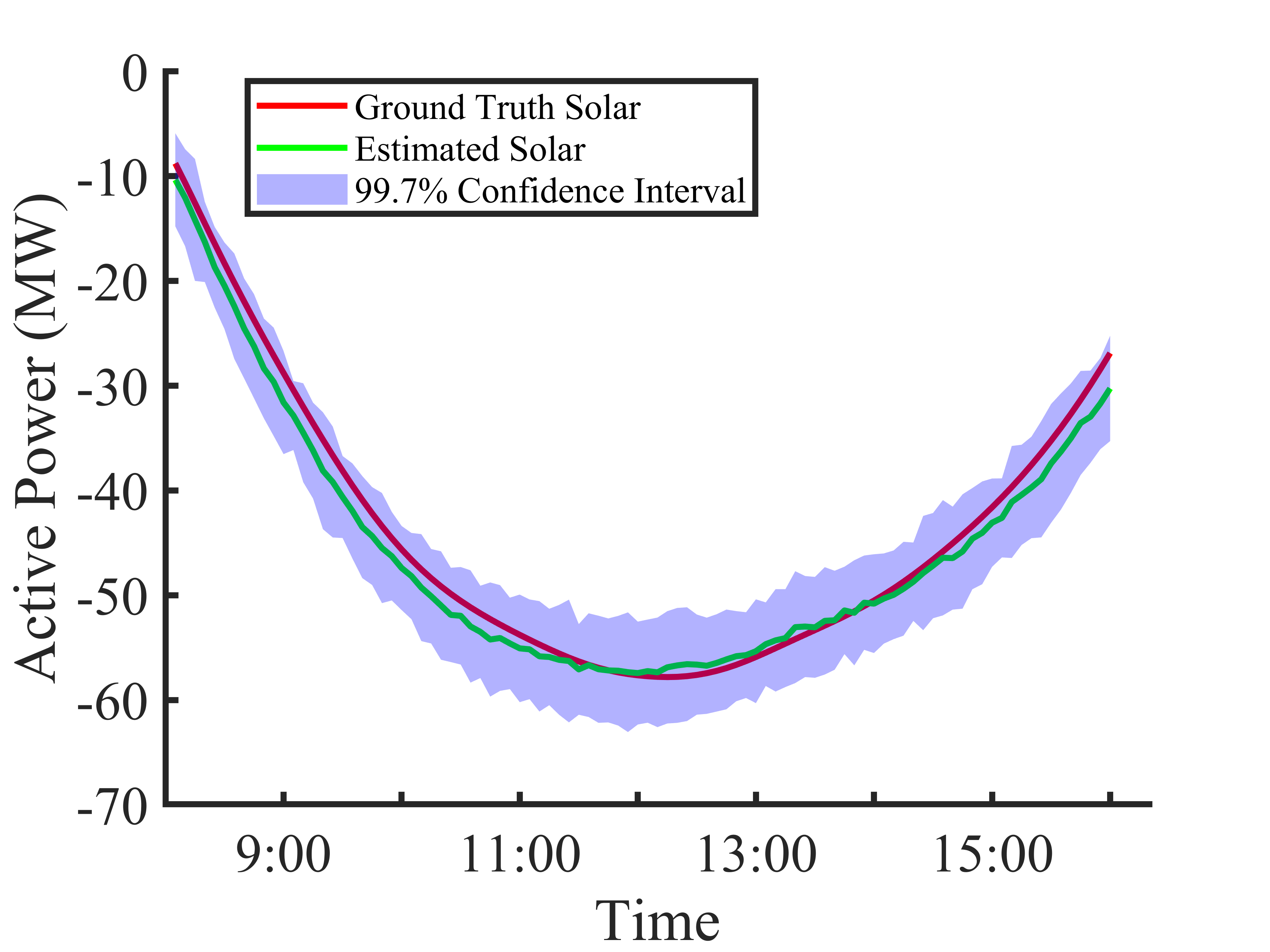}} 
      \addtocounter{subfigure}{+2}
    \subfigure[]{\includegraphics[width=0.18\textwidth]{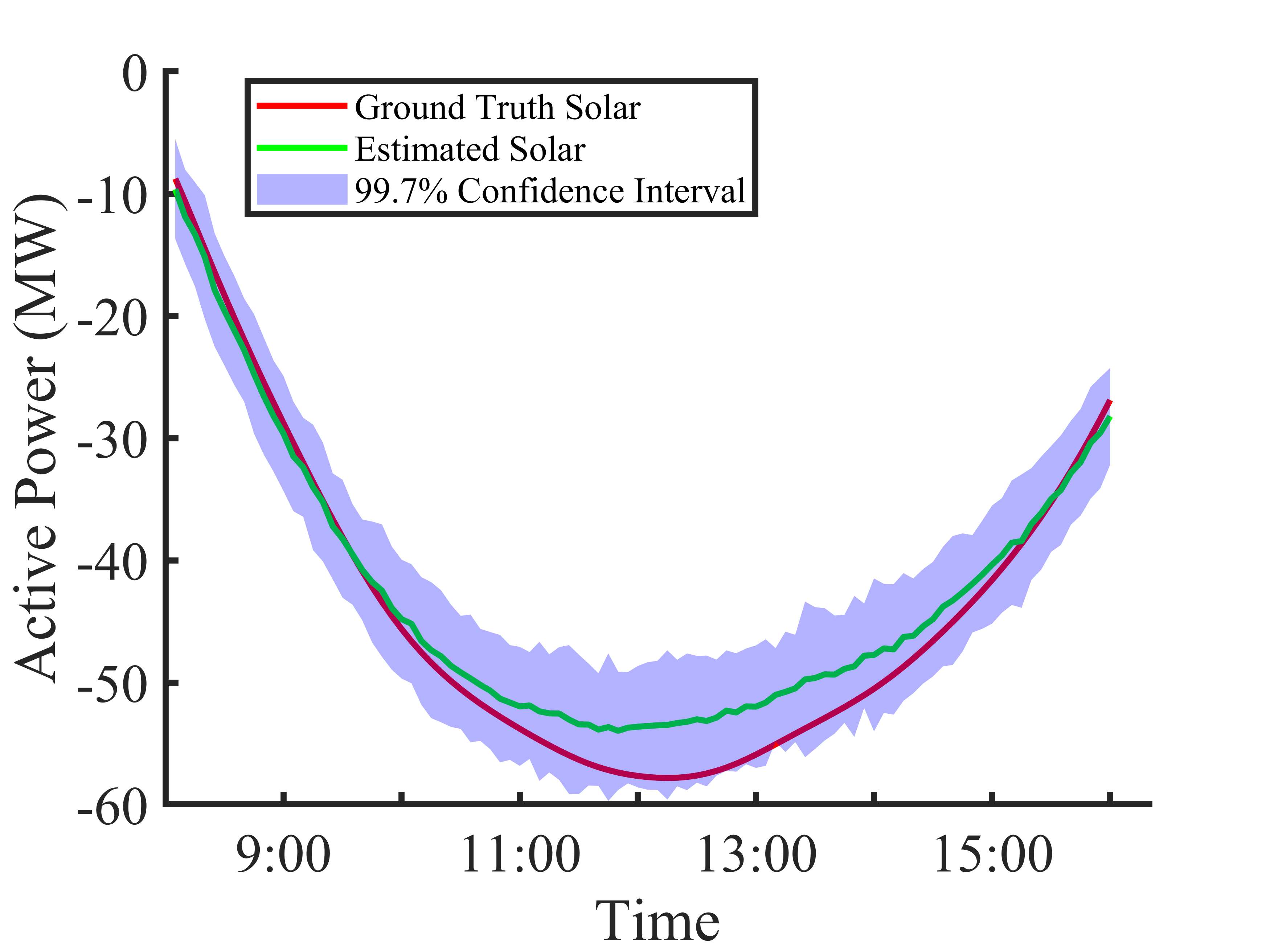}} 
          \addtocounter{subfigure}{+2}
    \subfigure[]{\includegraphics[width=0.18\textwidth]{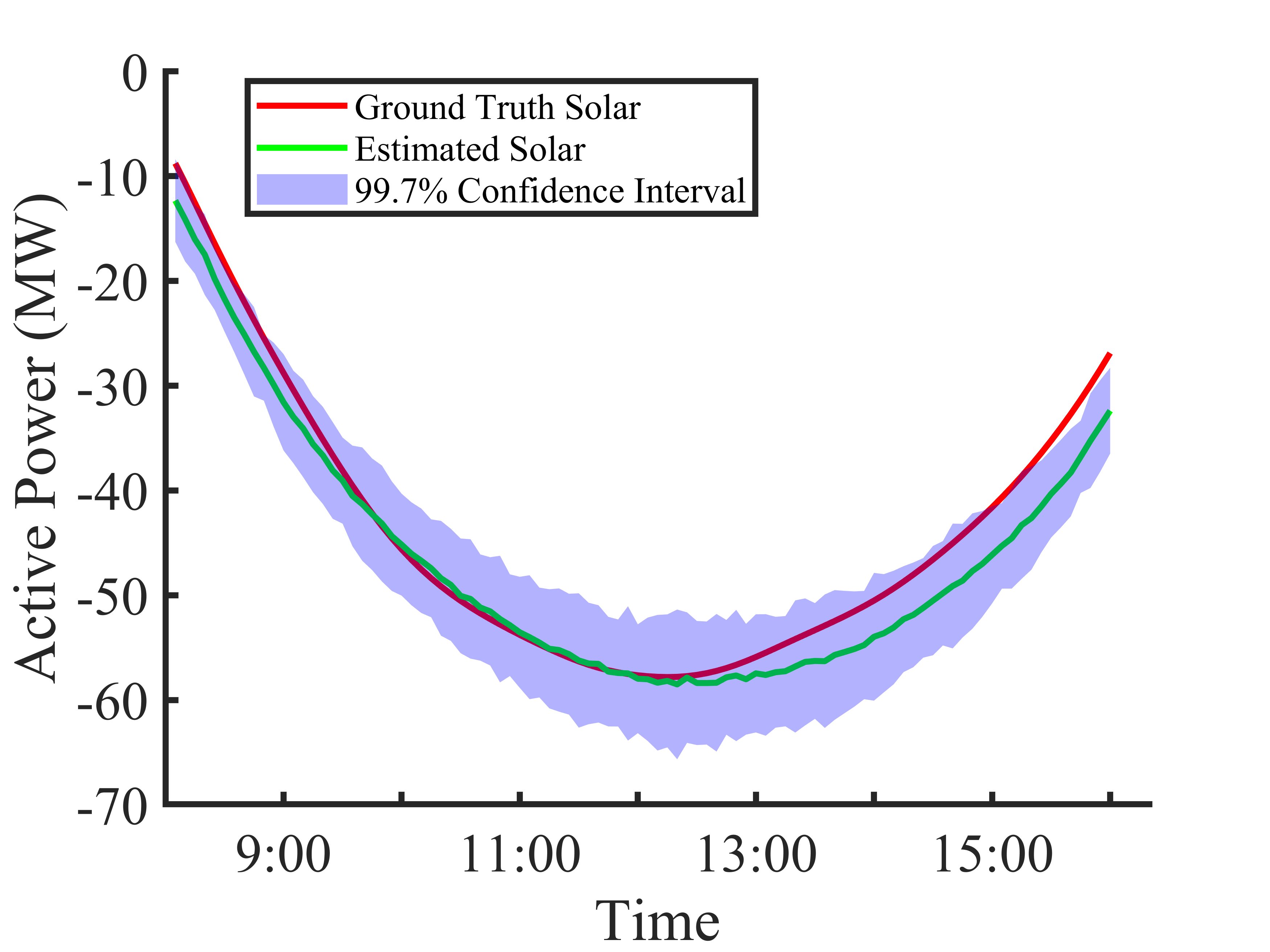}}
          \addtocounter{subfigure}{+2}
    \subfigure[]{\includegraphics[width=0.18\textwidth]{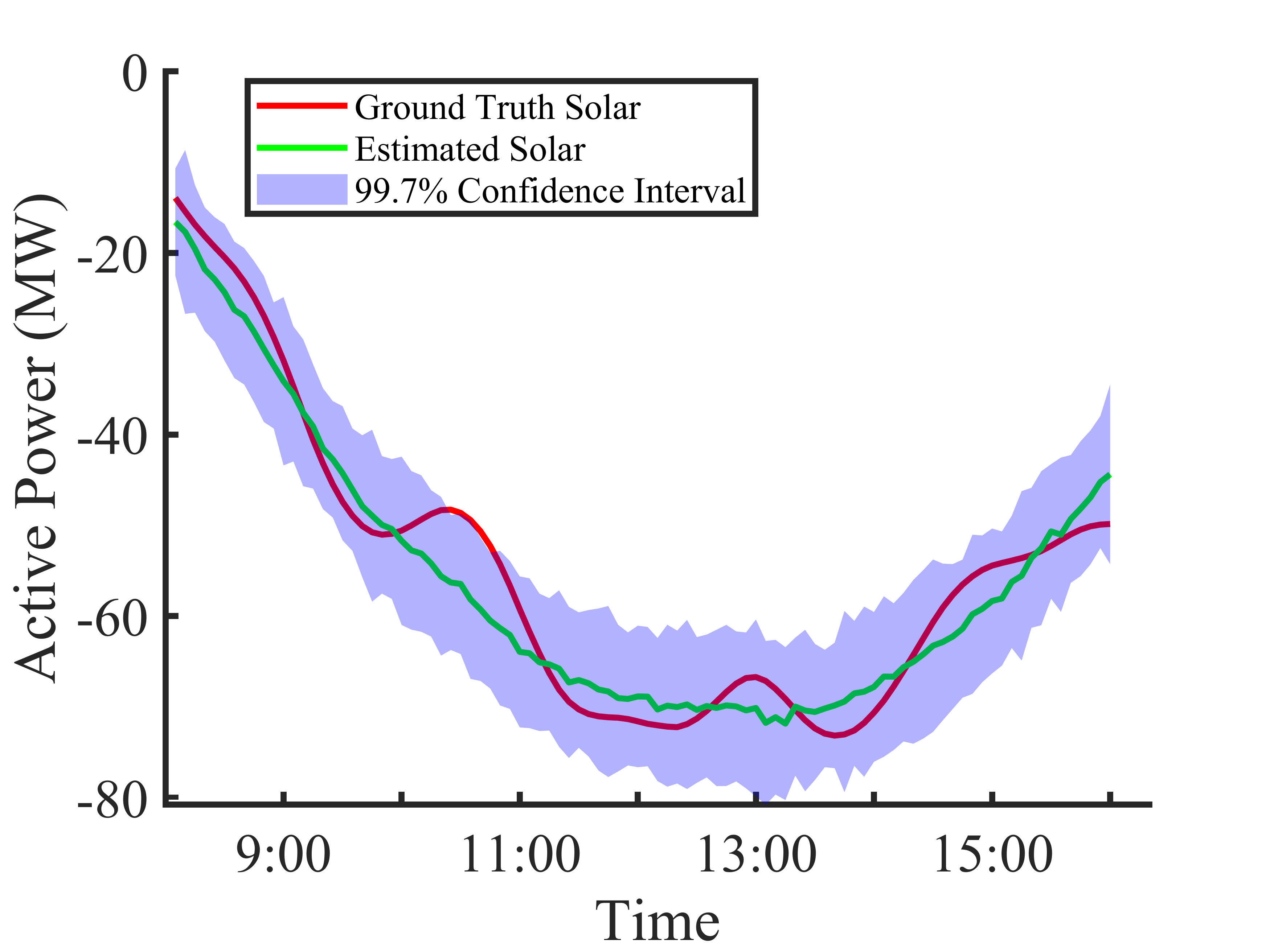}}
          \addtocounter{subfigure}{+2}
     \subfigure[]{\includegraphics[width=0.184\textwidth]{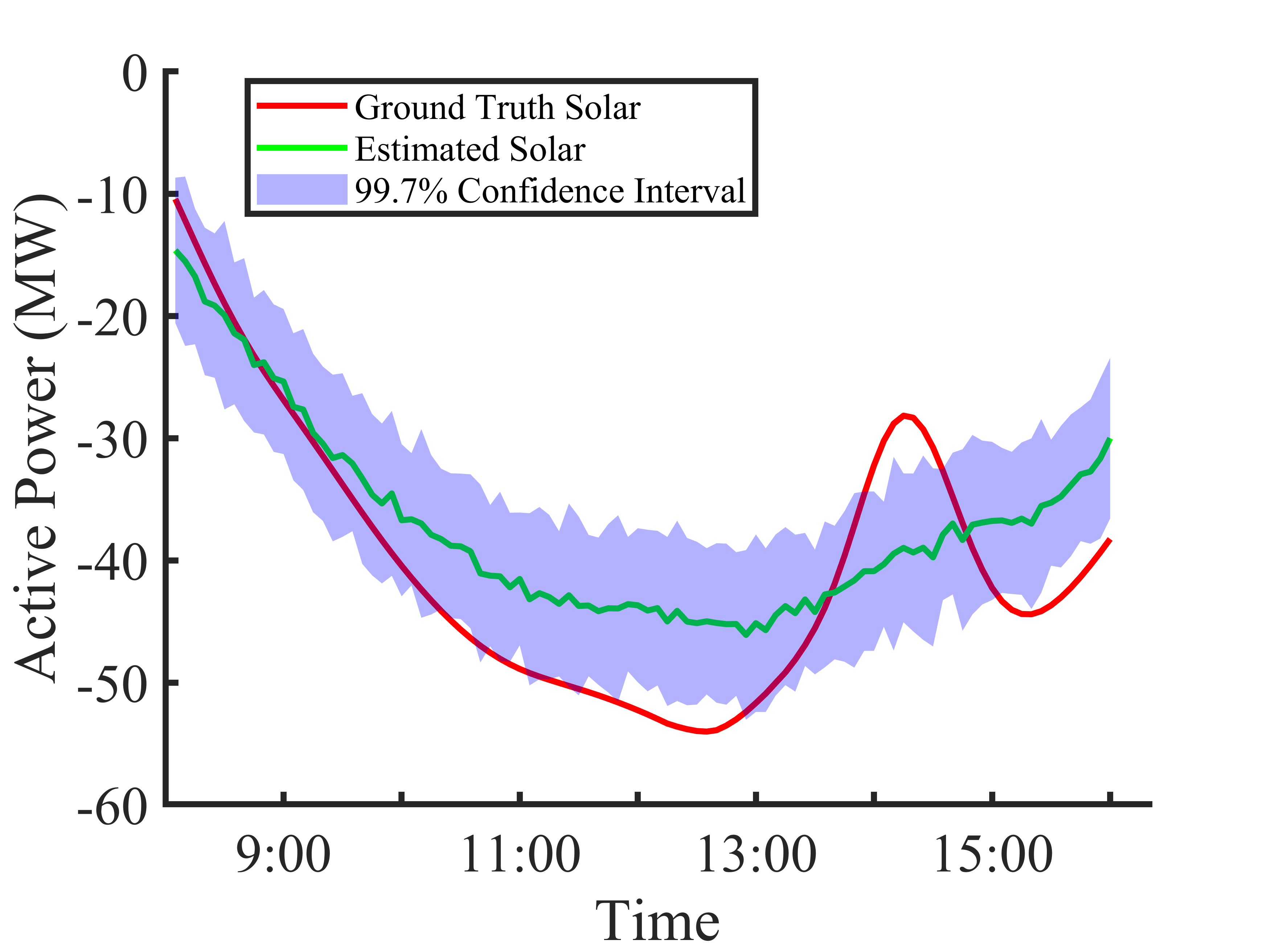}}
    	\caption{The disaggregation performance  of B-EDS on Cases 1-5. Each subfigure shows the ground-truth load, disaggregated load and the corresponding confidence interval. (a)-(c): the disaggregation results for Case 1.  (d)-(f): the disaggregation results for Case 2 where the test data contains Gaussian noise $\mathcal{N}(0,4^2)$. (g)-(i):  the disaggregation results for Case 2 where the test data contains Gaussian noise  $\mathcal{N}(0,6^2)$. (j)-(l):  the disaggregation results for Case 4 where the test data is a solar generation and its pattern  is different from the training data.  (m)-(o) the disaggregation results for Case 5 where the test data contains three loads, and the pattern of solar generation is different from the training data. }\label{fig_uncertainty}
\end{figure*}

\begin{table}[ht!]
\centering
\caption{The uncertainty indices and disaggregation accuracy on five testing cases}
\label{table_uncertainty}
\begin{tabular*}{\hsize}{@{}@{\extracolsep{\fill}}c|ccccc@{}}
\hline     & Case 1 & Case 2 & Case 3 & Case 4 & Case 5  \\
\hline U$_1$ &	243.72	&	280.35	&	201.52	&	0.058	&	160.89	\\
\hline U$_2$&	116.07	&	101.24	&	215.23	&	0.060	&	394.41	\\
\hline U$_3$ &	249.89	&	257.78	&	287.23	&	703.48	&440.26	\\
\hline U$_{\textrm{all}}$ &609.69	&	639.37	&	703.98	&	703.60	&788.44	\\
\hline B-EDS TER &	4.77\%	&	5.10\%	&	7.00\%	&	6.77\%	&	12.97\%	\\
\hline D-EDS TER &	7.19\%	&	8.86\%	&	11.60\%	&	11.01\%	&	16.45\%	\\
\hline 
\end{tabular*}
\end{table}

The Bayesian method B-EDS has slightly better disaggregation performance than 
the deterministic approach D-EDS.   The major advantage of the Bayesian approach is to measure the confidence level of the disaggregation results. However, the deterministic approach 
is much  more computationally efficient than the Bayesian method. {In Table I, the B-EDS requires around 50 seconds for the offline training  and 4 seconds for each testing sample. In comparison, the D-EDS requires around 15 seconds for the offline training, and 0.9 seconds for each testing sample. } If users want to know the reliability of the estimation results, the Bayesian method should be selected. In contrast, if users need to disaggregate the aggregate measurement with limited computational resources in real-time, the deterministic approach is a better option.

\section{Conclusion}
{


Energy disaggregation at substations with BTM solar generations has drawn increasing attention. 
Accurate energy disaggregation results are crucial for  power system planning and operations.  However, collecting training data with full labels at the substation level is challenging. Therefore, 
we propose the concept of partially labeled data which is   applicable in practice and significantly reduces the burden of annotating data. This paper summarizes two new  load disaggregation approaches. 
Both the deterministic approach and the Bayesian approach can achieve   accurate disaggregation results on partially labeled data. Moreover, 
an uncertainty index is proposed to measure the reliability of the disaggregation results. 
To the best of our knowledge, this is the first work to provide the uncertainty measure for the energy disaggregation problem.}


\section*{Acknowledgment}

This work was supported in part by the NSF grant \# 1932196, AFOSR FA9550-20-1-0122, and ARO W911NF-21-1-0255.



%

\bibliographystyle{IEEEtran} 
\bibliography{IEEEabrv,ref,MengWangPub}

\end{document}